\documentclass[conference]{IEEEtran}
\IEEEoverridecommandlockouts
\usepackage{cite}
\usepackage{amsmath,amssymb,amsfonts}
\usepackage{algorithmic}
\usepackage{graphicx}
\usepackage{textcomp}
\usepackage{xcolor}
\usepackage{url}
\usepackage{microtype}
\usepackage[hidelinks]{hyperref}
\usepackage{multirow}
\usepackage{array}
\usepackage{stfloats}
\usepackage{bm}
\usepackage{xspace}
\usepackage{enumitem}
\usepackage{booktabs}
\usepackage[mathscr]{eucal}
\usepackage{bbm}
\usepackage{makecell}
\usepackage{pifont}
\usepackage{float}
\usepackage{subfig}
\usepackage[export]{adjustbox}
\usepackage{balance}

\newcommand{\vpara}[1]{\vspace{0.05in}\noindent \textbf{#1 }}
\newcommand{\model}{OpenIMA}
\newcommand{\smodel}{OpenIMA\space}
\newtheorem{problem}{\textbf{Problem}}
\newtheorem{definition}{\textbf{Definition}}
\newtheorem{theorem}{\textbf{Theorem}}
\newtheorem{lemma}{\textbf{Lemma}}

\def\BibTeX{{\rm B\kern-.05em{\sc i\kern-.025em b}\kern-.08em
    T\kern-.1667em\lower.7ex\hbox{E}\kern-.125emX}}

\begin{document}

\title{Open-World Semi-Supervised Learning for Node Classification
}

\author{
	\IEEEauthorblockN{
		Yanling Wang$^{1,2}$, 
		Jing Zhang$^{1*}$, 
		Lingxi Zhang$^{1}$, 
		Lixin Liu$^{3}$,
            Yuxiao Dong$^{4}$\\
            Cuiping Li$^{1}$,
            Hong Chen$^{1}$,
	    Hongzhi Yin$^{5}$} 

        \IEEEauthorblockA{$^{1}$ {School of Information, Renmin University of China} $^{2}$ {Zhongguancun Laboratory}}
 
	\IEEEauthorblockA{$^{3}$ {Alibaba Group} $^{4}$ {Department of Computer Science and Technology, Tsinghua University}}
 
	\IEEEauthorblockA{$^{5}$ {School of Information Technology and Electrical Engineering, The University of Queensland}}

 \IEEEauthorblockA{wangyl@zgclab.edu.cn, \{zhang-jing, zhanglingxi, licuiping, chong\}@ruc.edu.cn, liulixin@pku.edu.cn
 \\yuxiaod@tsinghua.edu.cn, h.yin1@uq.edu.au}
 \thanks{$*$ Corresponding author}
} 

\maketitle

\begin{abstract}
Open-world semi-supervised learning (Open-world SSL) for node classification, that classifies unlabeled nodes into seen classes or multiple novel classes, is a practical but under-explored problem in the graph community.
As only seen classes have human labels, they are usually better learned than novel classes, and thus exhibit smaller intra-class variances within the embedding space (named as imbalance of intra-class variances between seen and novel classes). 
Based on empirical and theoretical analysis, we find the variance imbalance can negatively impact the model performance.
Pre-trained feature encoders can alleviate this issue via producing compact representations for novel classes. 
However, creating general pre-trained encoders for various types of graph data has been proven to be challenging. 
As such, there is a demand for an effective method that does not rely on pre-trained graph encoders.
In this paper, we propose an \textbf{IM}balance-\textbf{A}ware method named \smodel for \textbf{Open}-world semi-supervised node classification, which trains the node classification model from scratch via contrastive learning with bias-reduced pseudo labels.
Extensive experiments on seven popular graph benchmarks demonstrate the effectiveness of \model, and the source code has been available on GitHub\footnote{\href{https://github.com/RUCKBReasoning/OpenIMA}{https://github.com/RUCKBReasoning/OpenIMA}}.
\end{abstract}


\section{Introduction}\label{sec:intro}
\begin{figure}
    \centering
    \subfloat[\textmd{An example of open-world SSL in a coauthor network.
    }]{           
        \includegraphics[width=0.975\linewidth,valign=b]{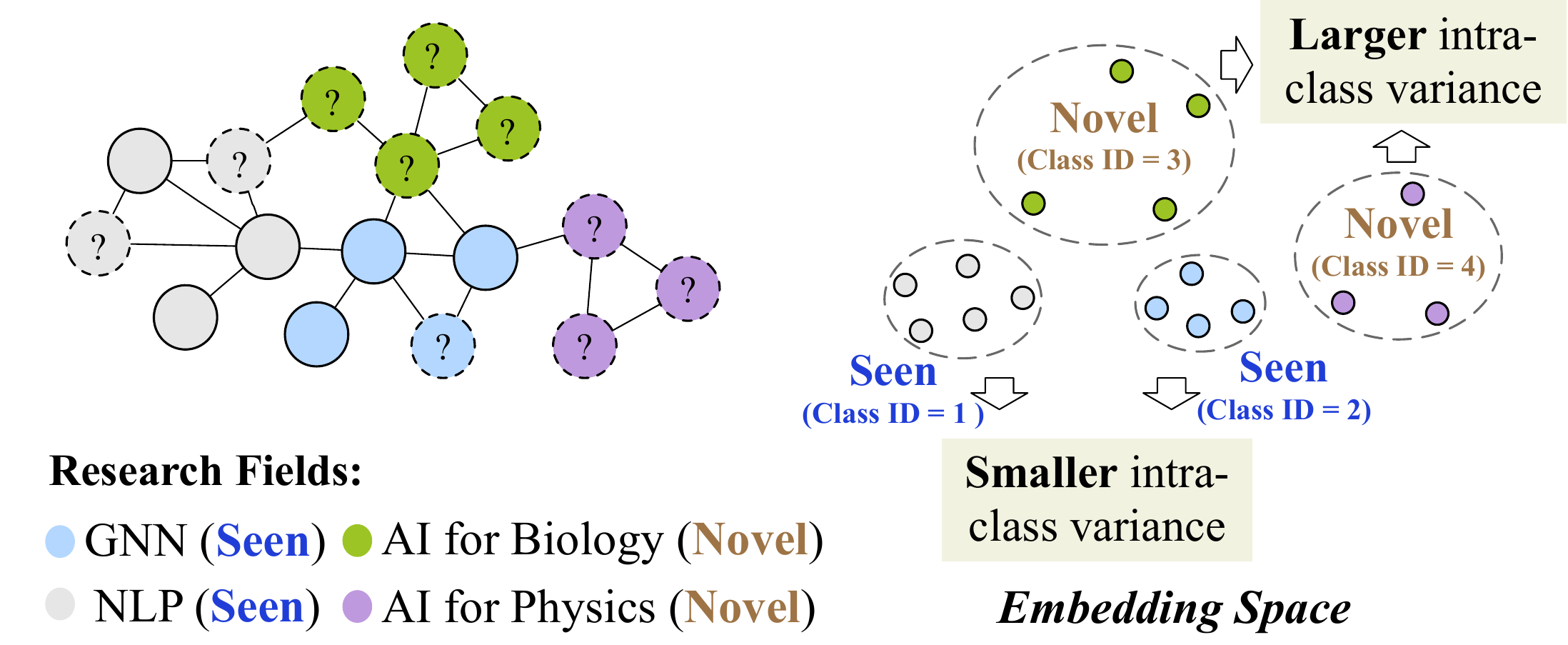} 
        \label{fig:GCD_example}          
        }
        \
        \
    \subfloat[\textmd{Effects of the variance imbalance on the Coauthor CS dataset. The imbalance rate quantifies the imbalance of intra-class variances between seen and novel classes, and the separation rate measures the separation between seen and novel classes. Detailed calculation method is introduced in Section~\ref{sec:appendix_imbalance_separation_rates}}]{
        \adjustbox{valign=b}{
        \small
        \renewcommand\tabcolsep{1.3pt}
        \small
\centering
  \renewcommand\arraystretch{0.85}
  \newcolumntype{?}{!{\vrule width 1pt}}
\begin{tabular}{l?cccc}
    \toprule
    &\multirow{2}*{\makecell[c]{Imbalance \\ Rate}}&\multirow{2}*{\makecell[c]{Separation \\ Rate}}&\multicolumn{2}{c}{Test Accuracy}\\
    \cline{4-5}
    &&&Seen& Novel\\
    \midrule
     InfoNCE& 1.002 & 1.239 & 0.728 & 0.727\\
     InfoNCE+SupCon & 1.071 & 1.271 & 0.751 $\uparrow$ & 0.710 $\downarrow$\\
     InfoNCE+SupCon+CE & 1.089 & 1.275 & 0.771 $\uparrow$ & 0.730 $\uparrow$\\
     \midrule
     \textbf{\smodel(Ours)} & 1.048 & 1.430 & 0.783 $\uparrow$ & 0.759 $\uparrow$\\
    \bottomrule
\end{tabular}

        \label{fig:motivation_table}}}
        
    \caption{Motivations of this work. \textmd{(a) The solid circles represent labeled nodes, and the dashed circles represent unlabeled nodes. (b) The results are averaged over ten runs. Model performance on other datasets are reported in Table~\ref{tab:overall_result}.}}
    \centering
    \label{fig:intro}
\end{figure}

With the advance of graph neural networks (GNNs), semi-supervised node classification~\cite{GCN,Graphsage,GIN,GeniePathAAAI,yanling2021DCI,ImbalancedNodeClf,ImGAGN} has achieved remarkable breakthroughs.
Despite the success, most node classification models are developed under the closed-world setting, assuming that both labeled and unlabeled data originate from the same set of classes. In real-world scenarios, this assumption rarely holds true because it is impractical for humans to label all possible classes in the world. 
Therefore, some studies~\cite{openWGL, OODGAT, lifelong1, lifelong2, iclr23@CIDER, icdm2022@DVOpenWorldNodeClf} explored open-world node classification, where test nodes are assumed to be from either seen classes or novel classes. These efforts treat nodes of novel classes as out-of-distribution (OOD) instances, thus assigning them into an outlier class.
More practically, our work aims to classify test nodes into previously seen classes or multiple novel classes, formally named as open-world semi-supervised learning (open-world SSL) for node classification.
Figure~\ref{fig:GCD_example} illustrates an example in a coauthor network, where nodes represent authors, edges depict coauthor relationships, and the colors of the nodes denote the authors' primary research fields.
Since novel research fields emerge over time, and it is challenging to promptly update the label information, a model is desired to classify unlabeled authors into previously seen fields or newly emerging fields.

The key to solving this problem lies in high-quality node representation learning.
However, as only seen classes have manually labeled nodes, the seen classes are often better learned than the novel classes to exhibit smaller intra-class variances in the embedding space. We illustrate the imbalance of intra-class variances between seen and novel classes on the right side of Figure~\ref{fig:GCD_example}.
In fact, the variance imbalance also occurs on other types of data.
To alleviate the issue, efforts~\cite{GCD, SimGCD,PromptCAL,MIB,LPS,GPC,ORCA,CPL} in the field of computer vision commonly use a pre-trained ResNet~\cite{resnet} or a pre-trained Vision Transformer (ViT)~\cite{ViT-B} to obtain more compact representations for samples of novel classes.
Nevertheless, graph community presently lacks general pre-trained feature encoders because data distributions can vary significantly across different graph types~\cite{GCC,graphormer}.
Although pre-trained graph encoders like GCC~\cite{GCC} and Graphormer~\cite{graphormer} have been proposed,
the related studies typically conduct pre-training and fine-tuning/prompting~\cite{AllinOne,GraphPrompt,GPPT} on the same or similar types of graph data, which limits their applicability.
Without a general and powerful pre-trained graph encoder, the variance imbalance between seen and novel classes naturally becomes a core factor to impact open-world SSL on a graph.
Motivated by this, we target to (1) explore how the variance imbalance affects open-world SSL for node classification, and (2) design an effective method to alleviate the variance imbalance issue without relying on pre-trained graph encoders.

To achieve the goal, we start with a preliminary experiment on the well-known Coauthor CS graph dataset~\cite{oleksandr@2018GNNEval}. Firstly, we divide classes into seen classes and novel classes, and construct a labeled set and an unlabeled set (Cf. Section~\ref{sec:exp}).  
To quantify the variance imbalance and measure the separation between seen and novel classes, we introduce two metrics --- imbalance rate and separation rate in Section~\ref{sec:appendix_imbalance_separation_rates}.
Initially, we employ the unsupervised contrastive loss InfoNCE~\cite{aaron2018InfoNCE} to learn representations for both labeled and unlabeled nodes, enabling unbiased learning of seen and novel classes.
Then we successively add supervised losses to retrain the model, including a supervised contrastive loss SupCon~\cite{prannay2020supcon} and the classic cross-entropy (CE), to enhance the representation learning of labeled nodes. By doing this, the intra-class variances of seen classes can gradually decrease, leading to increasing imbalance rates (Cf. the second column of Figure~\ref{fig:motivation_table}).
Meanwhile, the seen classes become more separated from the novel classes, leading to increasing separation rates (Cf. the third column of Figure~\ref{fig:motivation_table}).
Under different imbalance/separation rates, we run K-Means~\cite{kmeans} over node representations to cluster nodes, and align seen classes with specific clusters for final node classification. 
From Figure~\ref{fig:motivation_table}, we make the following observations:
\begin{itemize}[leftmargin=1em]
    \item When the separation rate is relatively low, the imbalance rate plays a crucial role in impacting the model performance on novel classes.
    Clearly, increasing the imbalance rate would decrease the accuracy on novel classes (Cf. InfoNCE vs. InfoNCE+SupCon), as dispersed node representations of novel classes are prone to be incorrectly clustered.
    \item When the seen classes are thoroughly learned and become well separated from the novel classes, the increase of the imbalance rate would not hurt the performance on novel classes (Cf. InfoNCE+SupCon vs. InfoNCE+SupCon+CE).
\end{itemize} 
Our theoretical analysis in Section~\ref{sec:theory} supports these findings.

Inspired by the insights, we propose an \textbf{IM}balance-\textbf{A}ware method named \smodel for \textbf{Open}-world semi-supervised node classification --- a simple and effective method to suppress the imbalance rate and increase the separation rate (demonstrated in Figure~\ref{fig:motivation_table}).
Specifically, we introduce pseudo labels to reduce the intra-class variances of novel classes and enhance the learning of seen classes.
Generally, previous work trains a classifier in a supervised or semi-supervised manner for pseudo-labeling~\cite{SimGCD,opencon,OpenLDN}. However, due to the absence of powerful pre-trained feature encoders for representing nodes, the node classifier tends to be seriously biased towards the seen classes. 
Therefore, we convert to unsupervised clustering to foster a more equitable pseudo-labeling.
Considering the unordered nature of cluster IDs, we align clusters with classes.
However, the clusters corresponding to novel classes can not be aligned, making the class IDs of novel classes  remain unordered.
Hence we adopt contrastive learning (CL) to learn the bias-reduced pseudo labels, as CL captures semantic
knowledge by learning whether two samples come from the same class instead of assigning them to specific classes as done in cross-entropy.

Overall, our work makes the following contributions:

\begin{itemize}[leftmargin=1em]
    \item Based on empirical and theoretical analysis, we find that the imbalance of intra-class variances between seen and novel classes is a critical factor that impacts open-world SSL on a graph. To our knowledge, this is a pioneer work to theoretically analyze the imbalance of intra-class variances between seen and novel classes. Moreover, we investigate open-world SSL for node classification, which has been less-explored before.   
    \item Based on our intriguing observations, we develop \smodel for open-world semi-supervised node classification, which 
    alleviates the variance imbalance issue to more effectively learn seen and novel classes from scratch.
    \item We conduct comprehensive experiments on seven real-world node classification datasets by comparing \smodel with representative baselines. The experimental results demonstrate the superiority of \smodel over the baselines. Especially on Coauthor Physics and ogbn-Products, \smodel derives 11.8\% and 12.5\% gains of overall accuracy compared with the best baseline, respectively.
\end{itemize}

\section{Related Work}\label{sec:related}
This work explores open-world SSL for node classification. Therefore, we first review research on open-world SSL. Then, we introduce tasks similar to open-world SSL and explain how open-world SSL differs from these tasks.
Finally, we review efforts on open-world node classification to explain differences between these works and \model, and then clarify the challenge of open-world SSL for node classification.

\subsection{Open-World Semi-Supervised Learning}
The goal of open-world SSL~\cite{GCD, SimGCD,PromptCAL,MIB,LPS,GPC,ORCA,CPL} is to group unlabeled instances into seen classes and multiple novel classes.
Considering that the novel classes lack human labels, existing open-world SSL methods usually perform pseudo-label generation to enable a self-training
style learning~\cite{opencon,OpenLDN,SimGCD,CPL}.
In general, a classifier composed of a feature encoder and a classification head, is trained to generate pseudo labels. However, as only seen classes have human labels, the classifier is prone to be biased towards the seen classes. 
In the field of computer vision, powerful pre-trained vision encoders are widely used to initialize the feature encoder~\cite{opencon,SimGCD,CPL}, which largely alleviates the difficulty in discriminating between seen and novel classes and thus reduces the biased prediction.
Nevertheless, when it comes to the graph community, due to the absence of general pre-trained graph encoders, we desire a more effective solution to generate bias-reduced pseudo labels.

For the variance imbalance between seen and novel classes, a previous work named ORCA~\cite{ORCA} has mentioned it as a critical factor to affect open-world SSL, while it does not theoretically analyze this factor. ORCA alleviates the variance imbalance by controlling the supervised learning of labeled data, so that the intra-class variances of seen classes can be similar to that of novel classes. Consequently, such a solution may suppress the performance on seen classes (Cf. ORCA-ZM vs. ORCA in Table~\ref{tab:overall_result}).
In contrast, our method targets to effectively learn both seen and novel classes.

\subsection{Open-World Semi-Supervised Learning Related Tasks}
Open-world SSL is related to but distinct from several open-world learning tasks, including open-set recognition (OSR)~\cite{OSR1, OSR2, OSR-Survey}, robust semi-supervised learning (robust SSL)~\cite{robust-SSL1,robust-SSL2,robust-SSL3, OpenMatch}, novel class discovery (NCD)~\cite{NCD-CCN,NCD-OpenMix,NCD-RankStats,openWGL}, and zero-shot learning (ZSL)~\cite{GZSL1,GZSL2,GZSL3}.
To distinguish between them, we divide them into three categories --- $C$+1 classification, $\bar{C}$ classification, and $C$+$\bar{C}$ classification (Cf. Table~\ref{tab:differences_open_world_node_classification_settings}).

OSR and robust SSL are $C$+1 classification tasks, where they regard instances of novel classes as OOD instances and perform $C$+1 classification to group the OOD instances into one single class.
Specially, OSR considers inductive setting, while robust SSL considers transductive setting.
NCD is a $\bar{C}$ classification task, as it classifies the OOD instances into multiple novel classes but assumes that the unlabeled data only involves novel classes.
Both open-world SSL and ZSL are $C$+$\bar{C}$ classification tasks. However, 
ZSL requires prior knowledge like semantic attributes of classes, so that it can learn mappings between samples and the prior class semantic information for prediction.
A recent approach called Zero-Knowledge ZSL (ZK-ZSL)~\cite{ZKZSL} does not use semantic attributes of novel classes, while it still requires the prior knowledge of seen classes. 
Compared to ZSL, open-world SSL is more flexible because it does not make any assumption about the prior knowledge of any class. 
Notably, $C$+1 classification methods can be adjusted for open-world SSL by post-clustering the detected OOD instances. Additionally, $\bar{C}$ classification methods can be adjusted for open-world SSL by aligning some of the discovered novel classes with previously seen classes.
However, these extensions have been demonstrated to not perform well~\cite{GCD,ORCA, OpenLDN}, indicating the necessity of specific methods for open-world SSL. 

\subsection{Open-World Node Classification}

Open-world node classification is an emerging task in the graph community.
Existing efforts primarily perform under the setting of $C$+1 classification~\cite{openWGL, OODGAT, lifelong1, lifelong2, icdm2022@DVOpenWorldNodeClf,G2Pxy}. For example, OODGAT~\cite{OODGAT} and OpenWGL~\cite{openWGL} treat nodes of novel classes as OOD instances and assign them into a single class.
In addition, open-world active learning has been explored on graph data~\cite{OWGAL}, while it requires iterative human annotation.
Different from previous work, this paper considers a more practical setting (i.e., open-world SSL), in which the OOD nodes will be grouped into multiple novel classes without human-in-the-loop.
Notably, a concurrent work devises an open-world SSL algorithm called RIGNN~\cite{luo2023rignn} for graph classification. However, it can not be used to solve the node classification problem. To our knowledge, \textit{\textbf{this paper is a pioneer work to explore open-world SSL for node classification and theoretically analyze factors that affect open-world SSL.}}

\textit{\textbf{Previous open-world SSL methods are primarily studied in the field of computer vision and thus often benefit from general pre-trained vision encoders. Due to an absence of general pre-trained graph encoders, these methods are sub-optimal when extended to graph data}} (Cf. Section~\ref{sec:exp}).
Concretely, 
open-world SSL for vision data~\cite{SimGCD,GCD,opencon, MIB} typically leverages a pre-trained vision encoder, such as a pre-trained ResNet~\cite{resnet} and a pre-trained Vision Transformer~\cite{ViT-B}, to initialize the feature encoder. The pre-trained encoders provide semantically meaningful initial representations, naturally facilitating discriminative representations for instances of both seen and novel classes. 
Nevertheless, there is an absence of general pre-trained graph encoders because different types of graphs have distinct data distributions (e.g., a social network and a biological network exhibit very different data properties).
Although pre-trained GNNs like GCC~\cite{GCC} and Graphormer~\cite{graphormer} have been proposed, they typically conduct pre-training and fine-tuning/prompting on the same or similar types of graphs, thus limiting their practical applications.
Therefore, in the absence of general pre-trained graph encoders, it is worth exploring more effective open-world SSL methods to train the node classification models from scratch.

\begin{table}
   \small
   \centering
   \renewcommand\arraystretch{0.95}
    \caption{Comparison between popular settings of open-world learning. $C$ denotes the number of seen classes, and $\bar{C}$ denotes the number of novel classes.}
    \newcolumntype{?}{!{\vrule width 1pt}}
    \setlength{\tabcolsep}{2.5mm}{
    \begin{tabular}{l?c?c}
        \toprule
        \textbf{Settings}& \textbf{Seen Classes} & \textbf{Novel Classes} \\
        \midrule
         $\bar{C}$ Classification & Not Classify & Classify\\
         $C$+1 Classification & Classify & Not Classify \\
         $C$+$\bar{C}$ Classification & Classify & Classify \\
        \bottomrule
    \end{tabular}
    }
    \label{tab:differences_open_world_node_classification_settings}
\end{table}

\section{Preliminary}\label{sec:problem}
\subsection{Problem Definition}
Let $G= \left(\mathcal{V}, \mathcal{E}, \mathcal{X}\right)$ be a graph, where $\mathcal{V}$ is the set of nodes, $\mathcal{E}$ is the set of edges, and $\mathcal{X}$ is the set of initial node features.
$\mathcal{V}$ can be divided into two distinct sets: a labeled set $\mathcal{V}_l$ and an unlabeled set $\mathcal{V}_u$. 
We denote the set of classes associated with the nodes in $\mathcal{V}_l$ as $\mathcal{C}_l$ (i.e., the set of seen classes) and the set of classes associated with the nodes in  $\mathcal{V}_u$ as $\mathcal{C}_u$.
In the open-world SSL problem, $\mathcal{C}_l \neq \mathcal{C}_u$ and $\mathcal{C}_l \cap \mathcal{C}_u \neq \emptyset$. The set of novel classes is $\mathcal{C}_n = \mathcal{C}_u \backslash \mathcal{C}_l$. 
\begin{problem}
	\textbf{Open-World SSL for Node Classification.} Given a partially labeled graph $G=\left(\mathcal{V}_l, \mathcal{V}_u, \mathcal{E}, \mathcal{X}, \mathcal{Y}_l\right)$, where $\mathcal{Y}_{l} = \{y_i | v_i \in \mathcal{V}_l, y_i \in \mathcal{C}_l\}$ is the set of manual labels.
    We abbreviate ``class label'' as ``label''.
 Let $\mathcal{Y}_u = \{y_i | v_i \in \mathcal{V}_u, y_i \in \mathcal{C}_u\}$ be the labels of $\mathcal{V}_u$, the goal is to learn a function
 
	\begin{equation}
		\mathcal{F}: {G} = \left(\mathcal{V}_l, \mathcal{V}_u, \mathcal{E}, \mathcal{X}, \mathcal{Y}_l\right) \rightarrow \mathcal{Y}_u
	\end{equation}
\end{problem}

\vpara{End-to-End Solution vs. Two-Stage Solution.} Existing efforts on open-world SSL can be categorized into end-to-end methods~\cite{OpenLDN, ORCA, opencon, MIB} and two-stage methods~\cite{GCD, PromptCAL}.
The former jointly optimizes a feature encoder and a classification head in an end-to-end manner, where the cross-entropy and an clustering objective~\cite{ORCA,OpenLDN,NCD-RankStats} are used together for model optimization. 
Due to supervised losses, end-to-end methods tend to over-fit seen classes, thereby needing well-designed regularizations and/or powerful pre-trained feature encoders for alleviation~\cite{ORCA,OpenLDN,opencon, MIB}.
Conversely, two-stage methods~\cite{GCD, PromptCAL} decouple representation learning and prediction, where they use non-parametric algorithms to cluster instances based on well-learned representations, and then align clusters with classes to obtain final prediction.
The two-stage methods have been demonstrated to reduce the bias. Inspired by this, we will follow the two-stage design.

\subsection{Calculation of Imbalance Rate and Separation Rate}\label{sec:appendix_imbalance_separation_rates}
Here we introduce how to calculate the imbalance rate and separation rate between seen and novel classes.
Given all the node representations of a class, we calculate the mean and standard deviation of these representations.
Given a seen class and a novel class, as well as their values of mean and standard deviation ($std$), we calculate the imbalance/separation rate by

\begin{equation}
    imbalance\_rate = \frac{\text{max}\left(std_{seen}, std_{novel}\right)}{ \text{min}\left(std_{seen}, std_{novel}\right)}
\end{equation}

\begin{equation}
    separation\_rate = \frac{||mean_{seen}-mean_{novel}||_2}{std_{seen}+ std_{novel}}
\end{equation}
The final imbalance rate and separation rate are averaged over all seen-novel class pairs.

\section{Methodology}\label{sec:method}
Based on intuitions from a theoretical model, we propose \smodel to perform open-world SSL for node classification.

\subsection{Theoretical Motivation} \label{sec:theory}
Variance imbalance (i.e., seen classes exhibit smaller intra-class variances than novel classes in the embedding space) is a potential key factor in affecting open-world SSL for node classification. Here we theoretically analyze how the variance imbalance affects model performance.
In specific, we consider $N$ data samples from a seen class and a novel class, and perform K-Means clustering.
Particularly, we assume the cluster IDs and class IDs are aligned, so
that the cluster labels can be used for classification.
In the theoretical model, the data-generation distribution $P_{XY}$ is a uniform mixture of two spherical Gaussian distributions. Please note $X$ here does not denote initial node features but denotes data samples in a latent feature space.
The class label $Y$ is either 1 or 2 with equal probability. Condition on $Y=1$, $X|Y \sim \mathcal{N}(\boldsymbol{\mu}_1,\Sigma_1)$, and $Y=2$, $X|Y \sim \mathcal{N}(\boldsymbol{\mu}_2,\Sigma_2)$.
Formally, K-Means ($K$=2) assigns each data sample to the nearest cluster by
\begin{equation}
    \hat{y} = \arg \min \limits_{j} ||\boldsymbol{x} - \boldsymbol{\theta}_j ||_2, j \in \{1, 2\}
\end{equation}
where $\boldsymbol{x}$ denotes a data sample, $\hat{y}$ denotes the predicted cluster label, and $\boldsymbol{\theta}_j$ denotes the center of the $j$-th cluster. 
As we assume that the cluster IDs and class IDs are aligned, we can use $\hat{y}$ to denote the predicted class label.
The cluster center is iteratively updated by the average of the data samples assigned to the cluster. 
It is straightforward to verify that the expectation of the converged cluster centers output by K-Means will be $\boldsymbol{\theta}_j = \mathbb{E}[X|\hat{Y}=j]$, and the expectation of clustering accuracy on each cluster will be $ACC_j = \mathbb{E}[\mathbbm{1}(\hat{Y}=j) | Y=j]$.

Let $\sigma_j^2$ denote the largest eigenvalue of $\Sigma_j$, i.e., $\sigma_j^2= {\lambda_{max}(\Sigma_j)}$.
A large value of $\sigma_j^2$ indicates that the corresponding class has a large intra-class variance.
We define the variance imbalance rate between the given two classes as $\gamma=\max(\sigma_1,\sigma_2)/\min(\sigma_1,\sigma_2)$.
Intuitively, as $\sigma_1$ and $\sigma_2$ are key parameters of $P_{XY}$, $\gamma$ could affect the clustering accuracy.
In particular, if the latent classes are separated with nearly no overlap, the cluster boundaries are easy to find, and thus K-Means can correctly partition data samples with high possibility regardless of the imbalance rate. 
We model this by the following theorem and prove it in Section~\ref{sec:proof}.

\begin{definition}
\label{def:separate}
Given two spherical Gaussian distributions $\mathcal{N}(\boldsymbol{\mu}_1,\Sigma_1)$ and $\mathcal{N}(\boldsymbol{\mu}_2,\Sigma_2)$, they are $\alpha$-separated if 
$$||\boldsymbol{\mu}_1-\boldsymbol{\mu}_2||_2 = \alpha (\sqrt{\lambda_{max}(\Sigma_1)}+\sqrt{\lambda_{max}(\Sigma_2))}) = \alpha\left(\sigma_1 + \sigma_2\right)$$

\noindent where $\alpha$ can be represented by $||\boldsymbol{\mu}_1-\boldsymbol{\mu}_2||_2 / \left(\sigma_1 + \sigma_2\right)$, so a large $\alpha$ indicates that the two distributions are highly separated.
\end{definition}

Assume that $\mathcal{N}(\boldsymbol{\mu}_1,\Sigma_1)$ and $\mathcal{N}(\boldsymbol{\mu}_2,\Sigma_2)$ are $\alpha$-separated. Without loss of generality, suppose that $\sigma_1 < \sigma_2$, we have: 

\begin{theorem}\label{theorem}
With $1<\gamma < 2$, for any $\delta$, there exists a constant $\overline{N}$, if the number of samples $N \geq \overline{N}$, with a possibility at least 1-$\delta$,\\
(1) if $1.5<\alpha<3$, $ACC_2$ and $\sigma_1$ a.s. are positively correlated;\\
(2) if $\alpha>3$, $|1- ACC_1|<0.05$ and $|1- ACC_2|<0.05$.
\end{theorem}

\vpara{Interpretations.} Let $\sigma_1$ be the intra-class variance of the seen class, and $\sigma_2$ be that of the novel class, we have $\sigma_1 < \sigma_2$.
The theorem illustrates two facts.
\begin{itemize}[leftmargin=1em]
    \item Increasing the imbalance rate can suppress the clustering accuracy on the novel class.
According to the first point of Theorem~\ref{theorem}, 
the decrease of $\sigma_1$ will lead to a decrease of $ACC_2$, while the imbalance rate $\gamma=\max(\sigma_1,\sigma_2)/\min(\sigma_1,\sigma_2)=\sigma_2 / \sigma_1$ will increase with the decrease of $\sigma_1$. Therefore, a negative correlation exists between $ACC_2$ and $\gamma$. In other words,
increasing the imbalance rate can adversely affect the clustering accuracy on the novel class.
    \item If seen and novel classes are well-separated in the embedding space, clustering accuracy on each class is hard to be affected by the imbalance rate.
Concretely, when the seen class is effectively learned and can be clearly distinguished from the novel class, we can say that 
the two classes are $\alpha$-separated, with $\alpha$ taking a large value. Based on the second point of Theorem~\ref{theorem}, the imbalance rate hardly affects the clustering results if $\alpha$ is large enough.
\end{itemize}

\vpara{Motivation.} Based on the above theoretical analysis, a model trained from scratch is desired to (1) reduce the intra-class variances of novel classes to get a lower imbalance rate, and (2) thoroughly learn the seen classes, encouraging their separation from novel classes. 

\subsection{Overview of \model}
Given a graph $G=\left(\mathcal{V}_l, \mathcal{V}_u, \mathcal{E}, \mathcal{X}, \mathcal{Y}_l\right)$, the inference of \smodel contains the following three steps:

\begin{itemize}[leftmargin=1em]
\item \textbf{Node embedding.}
A GNN encoder like GAT~\cite{GAT} encodes node representations $\mathcal{Z}_{l} = \{\boldsymbol{z}_i \in \mathbb{R}^{d} | v_i \in \mathcal{V}_l\}$ for labeled nodes and $\mathcal{Z}_{u} = \{\boldsymbol{z}_i \in \mathbb{R}^{d} | v_i \in \mathcal{V}_u\}$ for unlabeled nodes.

\item \textbf{Clustering.}
The classic algorithm K-Means~\cite{kmeans} is applied to cluster nodes based on their representations $\mathcal{Z}_{l} \cup \mathcal{Z}_{u}$, where the number of clusters is set to be $|\mathcal{C}_l|+|\mathcal{C}_n|$.
In fact, other clustering algorithms~\cite{tsung2021MiCE, junnan2021@PCL, vivek@2020CCL, mathilde@2020SwAV} can also be employed. 

\item \textbf{Cluster-Class Alignment and Node Classification.}
We run the Hungarian optimal assignment algorithm~\cite{hungarian} on the set of labeled nodes to find the optimal alignment between classes and clusters. Let $\mathcal{M}$ be the set of possible cluster-to-class mapping functions, the optimal alignment is found by
\begin{equation}
    m^{*} = \arg \max \limits_{m \in \mathcal{M}}\sum_{v_i \in \mathcal{V}_{l}}\mathbbm{1}\left\{y_i=m\left(o_i\right)\right\}    
\end{equation}
where $\mathbbm{1}$ is an indicator function which takes value of 1 if $y_i=m\left(o_i\right)$ and 0 otherwise. $o_i$ and $y_i$ are the cluster label and class label of the $i$-th node in $\mathcal{V}_l$, respectively.
With the optimal alignment $m^{*}$, we can predict class labels for the unlabeled nodes via $\hat{\mathcal{Y}}_u = m^{*}\left(\mathcal{O}_u\right) = \{m^{*}\left(o_i\right) | v_i \in \mathcal{V}_u\}$.
In particular, the clusters involve both seen and novel classes, so the number of clusters is more than that of seen classes, and thus a portion of cluster IDs will not be aligned.
\end{itemize}

\subsection{Objective Function} 
To optimize the GNN encoder, we employ the simple and effective cross-entropy (CE) to learn the valuable manual labels, and perform Contrastive Learning with Bias-reduced Pseudo labels (BPCL) to suppress the intra-class variances of novel classes and better separate seen classes from novel classes.

\begin{equation}
    \mathcal{L}_{\text{\model}} = \mathcal{L}_{\text{BPCL}} + \eta \mathcal{L}_{\text{CE}}
    \label{eq:GCD-BPCL}
\end{equation}

\noindent where $\eta$ is a scaling factor. 
Because we use the CE to optimize the GNN encoder, we introduce a classification head on top of the GNN encoder to derive logits for computing the CE loss.

The most popular method of pseudo-labeling is to train a classifier in a supervised/semi-supervised manner for pseudo label generation~\cite{LaSSL,SemiSupCLCocalibration,classAwareSemiSupCL,opencon,OpenLDN,CPL,SimGCD}.
However, only seen classes have manual labels, so the classifier tends to be biased towards seen classes (Cf. evaluation of OpenLDN/SimGCD in Table~\ref{tab:overall_result}).

\vpara{Bias-Reduced Pseudo Label Generation.}
In order to mitigate the above problem, we turn to unsupervised clustering, which is free from manual labels. Specifically, we run an unsupervised clustering algorithm over node representations $\mathcal{Z}_l \cup \mathcal{Z}_u$ to obtain cluster predictions $\mathcal{O}_l$ for labeled nodes and $\mathcal{O}_u$ for unlabeled nodes, where the number of clusters is $|\mathcal{C}_l|+|\mathcal{C}_n|$.
Intuitively, samples near the cluster centers are more likely to be correctly predicted. Hence we define the prediction confidence of node $v_i$ to be inversely proportional to 
$||\boldsymbol{z}_i - \boldsymbol{\theta}_{o_i}||_2$, where $o_i$ is the predicted cluster label of the node $v_i$, and $\boldsymbol{\theta}_{o_i}$ is the corresponding cluster center. 
According to the confidence values, we sort $\mathcal{O}_l \cup \mathcal{O}_u$ in descending order and select the top $\rho \%$ of them as the reliable ones to alleviate erroneous training.
As a portion of samples have been manually labeled, we only supplement pseudo labels $\mathcal{O}_u^{s} \subset \mathcal{O}_u$ for unlabeled nodes that exist in the top-$\rho\%$ set.
With the optimal cluster-to-class alignment $m^{*}$, we derive the final pseudo labels $\hat{\mathcal{Y}}_u^{s} = m^{*}\left(\mathcal{O}_u^{s}\right)$.
In this paper, we use the classic K-Means clustering. Other clustering algorithms~\cite{tsung2021MiCE, junnan2021@PCL, vivek@2020CCL, mathilde@2020SwAV} can also be employed. 

\vpara{Learning with Bias-Reduced Pseudo Labels.}
As cluster IDs are unordered, we need to align clusters and classes, so that we can use the manual labels and pseudo labels together.
However, different from the closed-world setting~\cite{M3S}, the cluster IDs corresponding to novel classes can not be aligned, making some class IDs remain unordered. 
Fortunately, contrastive learning (CL) can deal with unordered class IDs, as it captures semantic
knowledge by learning whether two samples come from the same class rather than assigning samples to specific classes like cross-entropy. Hence the presence of unordered class IDs will not hinder the pseudo label-enhanced learning.
In specific, we design the following CL scheme with our bias-reduced pseudo labels.

\textit{\textbf{Embedding-level CL.}}
Given $N_b$ randomly sampled nodes to form a mini-batch, we perform data augmentation on each sample twice to obtain positive pairs, and thus a batch contains $2N_b$ data points.
Let $\left(i, j\right)$ be the indices of a positive pair within the batch, we minimize the following objective,

\begin{equation}
    \small
    \mathcal{L}_{\text{BPCL}}^{emb}=-\frac{1}{2N_{b}}\sum_{i=1}^{2N_b} \frac{1}{|\mathcal{P}(i)|} \sum_{j \in \mathcal{P}(i)} \log \frac{\exp \left(\boldsymbol{z}_i^{\top} \cdot \boldsymbol{z}_j / \tau\right)}{\sum_{k=1}^{2N_{b}} \mathbbm{1}_{[k \neq i]} \exp \left(\boldsymbol{z}_i^{\top} \cdot \boldsymbol{z}_k / \tau\right)}
    \label{eq:BPCL_supcon_emb}
\end{equation}

\noindent where $\boldsymbol{z}_i$ denotes the $\mathcal{\ell}_2$-normalized representation of the $i$-th data point in the batch,
$\mathbbm{1}_{[k \neq i]}$ is the indicator function which takes value of 1 if $k \neq i$ and 0 otherwise, and $\tau$ is the temperature parameter.
Following SimCSE~\cite{danqi2021@SimCSE}, we pass the same input to the feature encoder twice. By using the standard dropout~\cite{dropout} twice, we can obtain the representations $\boldsymbol{z}_i$ and $\boldsymbol{z}_j$ as a ``positive pair''.
Within the batch, $\mathcal{P}(i)$ denotes the indices of the $i$-th data point's positive samples. The positive samples share the same class label with the $i$-th data point.
Notably, Eq.~\ref{eq:BPCL_supcon_emb} and SupCon~\cite{prannay2020supcon} have the same form. However, different from the fully supervised SupCon, BPCL defines positive pairs based on both manual labels and pseudo labels.
Specially for nodes without manual and pseudo labels, Eq.~\ref{eq:BPCL_supcon_emb} can be viewed as the InfoNCE loss because $|\mathcal{P}(i)|=1$.

\textit{\textbf{Logit-level CL.}}
We observe that combining CE loss can benefit the effectiveness of BPCL (Cf. Section~\ref{sec:ablation}). The CE loss is computed based logits output by the classification head. To better optimize the classification head, we propose to 
utilize the abundant pseudo labels of the unlabeled nodes. Considering that cross-entropy can not deal with the unordered class IDs of novel classes, we design the following logit-level objective,

\begin{equation}
    \small
     \mathcal{L}_{\text{BPCL}}^{logits}=-\frac{1}{2N_b}\sum_{i=1}^{2N_b} \frac{1}{|\mathcal{P}(i)|} \sum_{j \in \mathcal{P}(i)} \log \frac{\exp \left(\boldsymbol{e}_i^{\top} \cdot \boldsymbol{e}_j / \tau\right)}{\sum_{k=1}^{2N_b} \mathbbm{1}_{[k \neq i]} \exp \left(\boldsymbol{e}_i^{\top} \cdot \boldsymbol{e}_k / \tau\right)}
    \label{eq:BPCL_supcon_logits}
\end{equation}

\noindent where $\boldsymbol{e}_i = \text{l2\_norm}\left(W^{\top}\boldsymbol{z}_i\right)$ is the $\mathcal{\ell}_2$-normalized logits of the $i$-th data point in the batch, and $W$ is the weight matrix of the classification head (i.e., a linear layer). 

Finally, BPCL loss combines the above two CL objectives,
\begin{equation}
    \mathcal{L}_{\text{BPCL}} = \mathcal{L}_{\text{BPCL}}^{emb} + \mathcal{L}_{\text{BPCL}}^{logits}
    \label{eq:BPCL}
\end{equation}

\vpara{Discussions.}
Unlike some previous open-world SSL methods that adopt contrastive learning~\cite{GCD, ORCA, SimGCD,PromptCAL,MIB}, \smodel employs contrastive learning with the aim of better leveraging the bias-reduced pseudo labels.
Without the help of general pre-trained graph encoders, we mitigate the variance imbalance issue by learning with high-quality pseudo labels. Considering that our generated pseudo labels are bias-reduced but involve unordered class IDs, we employ contrastive learning to prevent unordered class IDs from hindering the learning process.
Please note that our goal is not to innovate a CL scheme but to propose an effective method for open-world SSL on a graph.

\vpara{Complexity Analysis of \model.}\label{sec:appendix_pseudocode}
{We denote the dimension of node representations as $d$, the number of nodes in a graph as $N$, the number of labeled nodes as $M$, the batch size as $N_b$, the number of clusters as $K$, and the number of iterations in K-Means as $T$. 
The time complexity can be divided into five parts: (1) The time complexity of CE-based supervised learning is $\mathcal{O}(MdK)$;
(2) The time complexity of BPCL is $\mathcal{O}(NdN_b)$; (3) The time complexity of K-Means is $\mathcal{O}(TNdK)$. Acceleration of K-Means has been widely studied~\cite{weizhong@2009parallelKmeans, gpuKmeans@2013you, minibatchKmeans}, which helps extend \smodel to large graphs.; (4) The time complexity of Hungarian algorithm is $\mathcal{O}(M^{3})$. If the size of labeled nodes is large, we can sample a subset of labeled nodes for computation; (5) The time complexity of selecting reliable pseudo labels is $\mathcal{O}(NlogN)$ as we need run a sorting algorithm.
Considering that $N_b$, $d$, $K$, $T$ and $M$ are relatively small, the complexity of \smodel is approximately about $O(NlogN)$.

\section{Experiments}\label{sec:exp}
\subsection{Experimental Settings}

\vpara{Datasets.}
The used datasets include two citation networks \textit{Citeseer}~\cite{GCN} and \textit{ogbn-Arxiv}~\cite{ogb}, three co-purchase networks \textit{Amazon Photos}~\cite{oleksandr@2018GNNEval}, \textit{Amazon Computers}~\cite{oleksandr@2018GNNEval}, and \textit{ogbn-Products}~\cite{ogb}, and two coauthor networks \textit{Coauthor CS}~\cite{oleksandr@2018GNNEval} and \textit{Coauthor Physics}~\cite{oleksandr@2018GNNEval}.
Statistics of them are in Table~\ref{tab:openima_datasets}.

\begin{table}
\small
\centering
	\caption{Statistics of the used datasets}
	\renewcommand\arraystretch{0.9}
	\begin{tabular}{@{}ccc@{~}c@{~}c@{}}
		\toprule
		\textbf{Graph} &\textbf{\#Nodes} &\textbf{\#Edges}& \textbf{\#Features} &\textbf{\#Classes} \\
		\midrule
		Citeseer&3,327&4,676&3,703&6\\
		Amazon Photos&7,650&119,082&745&8\\
		Amazon Computers&13,752&245,861&767&10\\
            Coauthor CS&18,333&818,94&6,805&15\\
            Coauthor Physics&34,493&247,962&8,415&5\\
            ogbn-Arxiv & 169,343 & 1,166,243 & 128 & 40\\
            ogbn-Products & 2,449,029 & 61,859,140 & 100 & 47\\
		\bottomrule
	\end{tabular}
	\label{tab:openima_datasets}
\end{table}

For each graph, we randomly select 50\% of classes as seen classes, and use the remaining classes as novel classes. 
For each seen class, we randomly sample 50 nodes (500 nodes for the larger datasets \textit{ogbn-Arxiv} and \textit{ogbn-Products}) to form the training set and another 50 nodes (500 nodes for \textit{ogbn-Arxiv} and \textit{ogbn-Products}) to form the validation set. The remaining nodes are assigned to the test set.
We use ten random seeds to derive ten train/validation/test splits for each graph.

\vpara{Baselines.}
We compare with two kinds of baselines. 
First, we compare with open-world node classification algorithms, showing that existing open-world node classification models which perform $C$+1 node classification are not suitable for extending to the open-world SSL setting. In specific, we compare with the typical algorithms --- \textbf{OODGAT}~\cite{OODGAT} and \textbf{OpenWGL}~\cite{openWGL}, where we cluster the detected out-of-distribution (OOD) nodes with K-Means to identify multiple novel classes.

Second, open-world SSL baselines are compared to show that they are sub-optimal for extending to the node classification, as they are originally proposed for vision data and often benefit from the powerful pre-trained feature encoder. Particularly, we replace the vision encoder with the graph encoder GAT~\cite{GAT}.
The end-to-end open-world SSL baselines include:
\begin{itemize}[leftmargin=1em]
\item \textbf{ORCA}~\cite{ORCA}, which designs an uncertainty adaptive margin mechanism to control the supervised learning of labeled data, so that intra-class variances of seen classes can be similar to that of novel classes, thus reducing the variance imbalance.
\item \textbf{ORCA-ZM}~\cite{ORCA}, which removes the margin mechanism from ORCA (i.e., ORCA with Zero Margin).
\item  \textbf{SimGCD}~\cite{SimGCD}, which trains a classifier with self-distillation and entropy regularization.
\item \textbf{OpenLDN}~\cite{OpenLDN}, which
trains a classifier with well-designed regularizer to generate pseudo labels, and learns the pseudo-labeled data using the standard cross-entropy loss.
\item \textbf{OpenCon}~\cite{opencon}, which trains a classifier to assign pseudo labels to the OOD samples and learns the pseudo labels of samples of novel classes via CL.
\end{itemize}

The two-stage open-world SSL baselines run K-Means over node representations, and align clusters with classes for node classification. The losses for representation learning include:
\begin{itemize}[leftmargin=1em]
\item \textbf{InfoNCE}~\cite{aaron2018InfoNCE}, which adopts the unsupervised CL loss InfoNCE to learn both labeled and unlabeled data. 
\item \textbf{InfoNCE+SupCon}, which additionally adds the supervised CL loss SupCon~\cite{prannay2020supcon} to learn labeled data.
\item \textbf{InfoNCE+SupCon+CE}, which further adds cross-entropy loss to enhance the learning of seen classes.
\end{itemize}

Notably, a semi-supervised K-Means algorithm is designed in GCD~\cite{GCD}, which forces samples of the same class to be clustered together. However, if a class has diverse node representations, such an operation could group samples of different classes together~\cite{yanling@2022ClusterSCL}. We find that K-Means performs better than the semi-supervised K-Means on the used datasets, so we adopt the classic and effective K-Means for clustering.

\vpara{Evaluation Metric.}
We adopt the widely used clustering accuracy~\cite{GCD,OpenLDN,ORCA,NCD-RankStats} for evaluation.
Given the ground-truth class labels and the predicted cluster labels of the test nodes,
we run the Hungarian algorithm to align the class IDs and cluster IDs.
After that, we can calculate the overall accuracy.
Following GCD~\cite{GCD}, we run the Hungarian assignment only once across all classes, and calculate the resultant accuracy on seen and novel classes, respectively.
We repeat all the experiments ten times with ten different train/validation/test splits, and the reported accuracy is averaged over the ten runs.

\begin{table*}
  \centering
  \renewcommand\arraystretch{0.85}
  \newcolumntype{?}{!{\vrule width 1pt}}
  \small
  \caption{Overall evaluation by test accuracy (\%). $^\dag$ indicates the $C$+1 node classification is extended to solve the open-world SSL setting by clustering the detected OOD nodes. $^\ddag$ denotes the two-stage variant of the original method. The bold numbers represent the best results, and the underlined numbers represent the second best results.}
    \label{tab:overall_result}
  \setlength{\tabcolsep}{1mm}{
    \begin{tabular}{p{3cm}?ccc?ccc?ccc?ccc?ccc}
    \toprule
     &\multicolumn{3}{c?}{\textbf{Citeseer}} & \multicolumn{3}{c?}{\makecell[c]{\textbf{Amazon}\\\textbf{Photos}}} & \multicolumn{3}{c?}{\makecell[c]{\textbf{Amazon}\\\textbf{Computers}}}&\multicolumn{3}{c?}{\makecell[c]{\textbf{Coauthor}\\\textbf{CS}}}
     &\multicolumn{3}{c}{\makecell[c]{\textbf{Coauthor}\\\textbf{Physics}}}\\   
    \cmidrule{2-16}
   & All & Seen & Novel & All & Seen & Novel & All & Seen & Novel & All & Seen & Novel & All & Seen & Novel \\
    \midrule 
    {OODGAT$^{\dag}$} & 46.4 & 56.9 & 37.5
                      & 63.0 & 71.1 & 54.5
                      & 61.3 & 63.3 & 55.9
                      & 68.1 & 68.8 & 65.6
                      & 68.3 & 69.4 & 62.5\\
    {OpenWGL$^{\dag}$} & 62.4 & 71.0 & 54.2
                       & 71.8 & 74.8 & 69.3
                       & 57.6 & 65.9 & 44.6
                       & 58.6 & 67.1 & 50.3
                       & \underline{73.3} & 85.0 & 68.1\\
    \midrule
    {ORCA-ZM} & 58.3 & 72.8 & 44.4
              & 74.6 & 89.9 & 58.2
              & \underline{63.8} & 73.7 & 52.6
              & \underline{75.0} & 74.2 & 73.5
              & 64.7 & 81.1 & 55.9\\
    {ORCA} & 58.2 & 68.0 & 49.0
           & 76.2 & 87.1 & 64.9
           & 60.9 & 67.8 & 53.7
           & 73.9 & 81.6 & 68.3
           & 66.2 & 84.8 & 58.2\\
    {SimGCD} & 61.5 & 70.6 & 53.4
             & 80.5 & 90.0 & 70.8
             & 61.9 & 73.8 & 50.3
             & 71.2 & 84.2 & 61.2
             & 60.9 & 81.1 & 52.8\\
    {OpenLDN} & 62.3 & 73.9 & 51.6 
              & 80.9 & 90.6 & 71.9
              & 63.3 & 76.5 & 51.8
              & 68.4 & 80.6 & 60.3
              & 62.2 & 72.4 & 57.2\\
    {OpenCon} & \textbf{68.8} & 75.0 & 62.1
              & 82.6 & 92.1 & 72.8
              & 62.3 & 74.9 & 51.2
              & 73.5 & 83.4 & 67.5
              & 65.8 & 95.0 & 55.4\\
    {OpenCon$^{\ddag}$} & 66.7 & 73.7 & 60.0
                         & \underline{82.9} & 87.9 & 78.1
                     & 59.4 & 69.0 & 53.2 
                     & 71.0 & 81.9 & 64.8
                     & 62.6 & 83.8 & 54.4\\
    \midrule
    {InfoNCE} & \underline{68.1} & 70.7 & 65.2
              & 76.3 & 78.5 & 75.1
              & 56.1 & 51.3 & 59.1
              & 72.2 & 72.8 & 72.7
              & 60.6 & 58.1 & 60.2\\
    {InfoNCE+SupCon} & \underline{68.1} & 71.9 & 64.1
                     & 75.6 & 80.3 & 72.0
                     & 56.3 & 52.5 & 58.9
                     & 72.4 & 75.1 & 71.0
                     & 60.5 & 59.7 & 59.8\\
    {InfoNCE+SupCon+CE} & \underline{68.1} & 73.6 & 62.6
                        & 76.4 & 80.5 & 72.9
                        & 55.8 & 54.7 & 56.5
                        & 74.4 & 77.1 & 73.0
                        & 62.8 & 79.4 & 56.1\\
    \midrule
    \textbf{\smodel}& \underline{68.1} & 71.8 & 64.3
                          & \textbf{83.6} & 89.9 & 77.3
                          & \textbf{67.8} & 77.8 & 59.0
                          & \textbf{77.1} & 78.3 & 75.9
                          & \textbf{78.0} & 93.6 & 72.2\\
    \bottomrule
   
    \end{tabular}
    }
\end{table*}

\vpara{Parameter Settings.}\label{sec:parameter_settings_sc&acc}
Detailed settings are introduced in Section~\ref{sec:appendix_hyperparam_setting}.
Here we introduce the metric for hyper-parameter selection.
For the closed-world problems, accuracy on the validation set is widely used for hyper-parameter selection.
However, in the open-world SSL problem, relying on the validation accuracy can make models biased to seen classes because the validation set is composed of only seen classes.
Recently, OpenCon~\cite{opencon} designs a validation strategy that splits the labeled classes into two parts, i.e., ``seen classes'' and ``novel classes'', to construct a virtual open-world SSL task.
The best hyper-parameters are then selected according to performance on the constructed task. 
However, the data distribution of the virtual task may significantly differ from that of the real-world scenario.
Moreover, it is hard to construct an effective validation task if the number of labeled classes is small.

Therefore, \textit{\textbf{we devise a metric called SC\&ACC for hyper-parameter selection under the open-world SSL setting}}, involving two popular metrics --- \textbf{S}ilhouette \textbf{C}oefficient (SC)~\cite{peter1987silhouettes} and Clustering \textbf{Acc}uracy (ACC). SC is a clustering metric that measures the clustering quality based on the representations and the predicted cluster labels.
SC\&ACC takes into account model performance on both seen and novel classes, thus alleviating the bias towards seen classes.
As the open-world SSL is a transductive problem where unlabeled data is available during training, we calculate ACC on the validation set and SC on the union of validation and test sets.
Under different combinations of hyper-parameters, we derive different (SC, ACC) value pairs.
We normalize the values of SC via min-max normalization and also perform min-max normalization for values of ACC. Then, for each combination of hyper-parameters, we take the weighted sum of the normalized SC and the normalized ACC (the weight takes value 0.5) to obtain the value of SC\&ACC. 

\subsection{Overall Evaluation}\label{exp:overall_eval}
From Table~\ref{tab:overall_result} and Table~\ref{tab:evaluation_large}, we observe that: 
\textit{\textbf{(1) Existing open-world node classification methods  are not suitable for
extending to the open-world SSL setting.}} \smodel outperforms OODGAT$^\dag$ and OpenWGL$^\dag$ on all the datastes.
\textit{\textbf{(2) Existing open-world SSL methods are sub-optimal when extended to the node classification setting}}.
Without powerful pre-trained feature encoders, the open-world SSL baselines can not guarantee good performance on both seen and novel classes.
\textit{\textbf{(3) \smodel can alleviate the variance imbalance issue.}}
Compared to InfoNCE which guides unbiased learning, \smodel not only achieves higher accuracy on seen classes but also derives better/comparable performance on novel classes.
\textit{\textbf{(4) \smodel better mitigates the variance imbalance issue compared to ORCA.}}
ORCA-ZM can outperform ORCA on seen classes, implying that controlling the learning of seen classes could suppress the learning quality. In contrast, \smodel can achieve good performance on both seen and novel classes.
\textit{\textbf{(5) The proposed bias-reduced pseudo-labeling is effective.}}
Compared to OpenLDN, SimGCD, and OpenCon, which train a classifier in a semi-supervised manner for pseudo-labeling, \smodel better balances the performance on seen and novel classes to derive better overall accuracy. 
Particularly, SimGCD and OpenCon can perform well on vision datasets with the assistance of powerful pre-trained vision encoders~\cite{SimGCD,opencon}.
\textit{\textbf{(6) Representations learned by the end-to-end methods are insufficient.}} We run K-Means over node representations learned by the competitive baseline OpenCon (denoted as OpenCon$^{\ddag}$). However, such a two-stage variant reduces overall accuracy on most of the datasets.
\textit{\textbf{(7) \smodel also performs well on larger datasets.}} We use mini batch-based K-Means~\cite{minibatchKmeans} for the larger datasets. Consequently, the clustering quality could be affected. Hence we refine \smodel for larger datasets via 1) predicting with the classification head and 2) adding a widely used pair-wise loss~\cite{ORCA} to mitigate the over-fitting of seen classes. 
As shown in Table~\ref{tab:evaluation_large}, \smodel still obtains the best overall accuracy compared with the competitive baselines. 

\begin{table}
   \small
   \centering
   \renewcommand\arraystretch{0.8}
    \caption{Evaluation on larger datasets by test accuracy (\%).}
    \newcolumntype{?}{!{\vrule width 1pt}}
    \setlength{\tabcolsep}{1.5mm}{
    \begin{tabular}{l?ccc?ccc}
        \toprule
        &\multicolumn{3}{c?}{\makecell[c]{\textbf{ogbn-Arxiv} \\ \textbf{(169,343 nodes)}}} & \multicolumn{3}{c}{\makecell[c]{\textbf{ogbn-Products} \\ \textbf{(2,449,029 nodes)}}} \\
        \cmidrule{2-7}
        & All &  Seen & Novel & All &  Seen & Novel \\
        \midrule
        ORCA-ZM& \underline{41.6} & 47.0 & 31.6 & \underline{49.5} & 61.5 & 32.3\\
        ORCA & \underline{41.6} & 44.7 & 34.6 & 46.8 & 55.5 & 34.3\\
        OpenCon & 32.2 & 31.8 & 31.6 & 43.7 & 46.0 & 43.0\\
        \midrule
         \textbf{\model} & \textbf{43.6} & 49.2 & 32.9 & \textbf{62.0} & 73.6 & 44.3\\
        \bottomrule
    \end{tabular}
    }
    \label{tab:evaluation_large}
\end{table}

\begin{table}
  \small
  \centering
  \renewcommand\arraystretch{0.8}
  \newcolumntype{?}{!{\vrule width 1pt}}
  \caption{Ablation studies by overall test accuracy (\%).}
    \label{tab:openima_ablation}
  \setlength{\tabcolsep}{0.15mm}{
    \begin{tabular}{lcc?c?c?c?c?c}
    \toprule
     \multirow{1}*{$\mathcal{L}_{\text{BPCL}}^{emb}$}&\multirow{1}*{$\mathcal{L}_{\text{BPCL}}^{logit}$}&\multirow{1}*{$\mathcal{L}_{\text{CE}}$}
     &{\textbf{Citeseer}} & {\makecell[c]{\textbf{Amazon}\\\textbf{Photos}}} & {\makecell[c]{\textbf{Amazon}\\\textbf{Computers}}} & {\makecell[c]{\textbf{Coauthor}\\\textbf{CS}}}
     &{\makecell[c]{\textbf{Coauthor}\\\textbf{Physics}}}\\  
    \midrule
    \ding{55} & \ding{55} & \ding{51} & 49.5 & 60.1 & 60.1 & 65.9 & 49.3\\

    \ding{51} & \ding{51} & \ding{55} & 67.8
                                      & 80.8
                                      & 55.8 
                                      & 76.0
                                      & 58.8 \\
                                      
    \midrule

    \ding{55} & \ding{51} & \ding{55} & 67.2 
                                      & 79.7
                                      & 56.5 
                                      & 73.4
                                      & 54.6 \\

    \ding{55} & \ding{51} & \ding{51} & 67.0  
                                      & 81.9 
                                      & 67.7
                                      & 75.8
                                      & \textbf{82.5} \\
    \midrule
                                      
    \ding{51} & \ding{55} & \ding{55} & 68.7
                                      & 80.6
                                      & 55.7 
                                      & 77.0  
                                      & 59.1 \\
                                      
    \ding{51} & \ding{55} & \ding{51} & \textbf{69.0}
                                      & 82.8
                                      & 66.4
                                      & \textbf{78.1}
                                      & 64.0
                                      \\
    \midrule
    
    \ding{51} & \ding{51} & \ding{51} & 68.1 
                                      & \textbf{83.6}  
                                      & \textbf{67.8}
                                      & 77.1
                                      & 78.0 \\
    \midrule
    \midrule
    \multicolumn{3}{c?}{Ours w/o PL} & 67.2 & 77.2 & 57.3 & 71.6 & 64.1 \\
     \bottomrule
    \end{tabular}
    }
\end{table}

\subsection{Ablation Studies}\label{sec:ablation}
We analyze components of $\mathcal{L}_{\text{\model}}$ according to Table~\ref{tab:openima_ablation}.
(1) $\text{BPCL}_{logit}$+$\text{BPCL}_{emb}$ performs better than $\text{CE}$ on most datasets because a large number of unlabeled data is not learned by CE. 
(2) $\text{BPCL}_{logit}$+$\text{CE}$ performs better than $\text{BPCL}_{logit}$ on four out of five datasets, and $\text{BPCL}_{emb}$+$\text{CE}$ performs better than $\text{BPCL}_{emb}$ on all the datasets, indicating that CE contributes to the BPCL.
(3) CE contributes most on Amazon Computers and Coauthor CS, and $\text{BPCL}_{logit}$+$\text{CE}$ performs better than $\text{BPCL}_{emb}$+$\text{CE}$ on the two datasets. CE loss is computed based on the classification head.
That is to say, if it is highly necessary to well optimize the classification head, logit-level BPCL takes effect as it brings rich information from the unlabeled data to enhance the learning of the classification head.
(4) $\text{BPCL}_{logit}$+$\text{CE}$ sometimes performs better than $\text{BPCL}_{emb}$+$\text{CE}$, while $\text{BPCL}_{emb}$+$\text{CE}$ sometimes performs better $\text{BPCL}_{logit}$+$\text{CE}$. {\smodel combines CE, $\text{BPCL}_{emb}$, and $\text{BPCL}_{logit}$ to derive satisfied performance across all the datasets}. (5) Removing bias-reduced pesudo-labeling from \smodel (i.e., ours w/o PL) decreases the model performance, demonstrating the effectiveness of our pseudo-labeling method.

\begin{figure}
    \centering
    \includegraphics[width=0.205\textwidth]{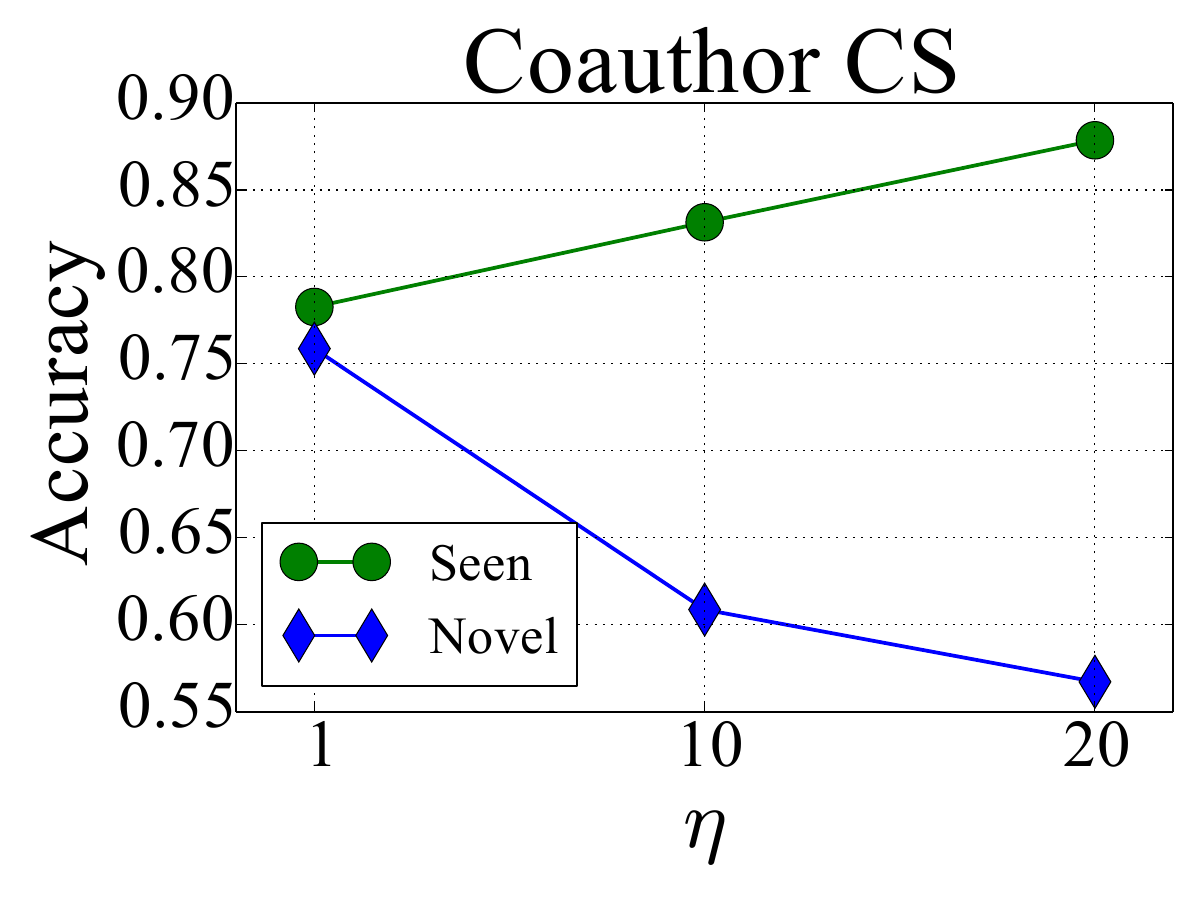}
    \hspace{0.05in}
    \includegraphics[width=0.205\textwidth]{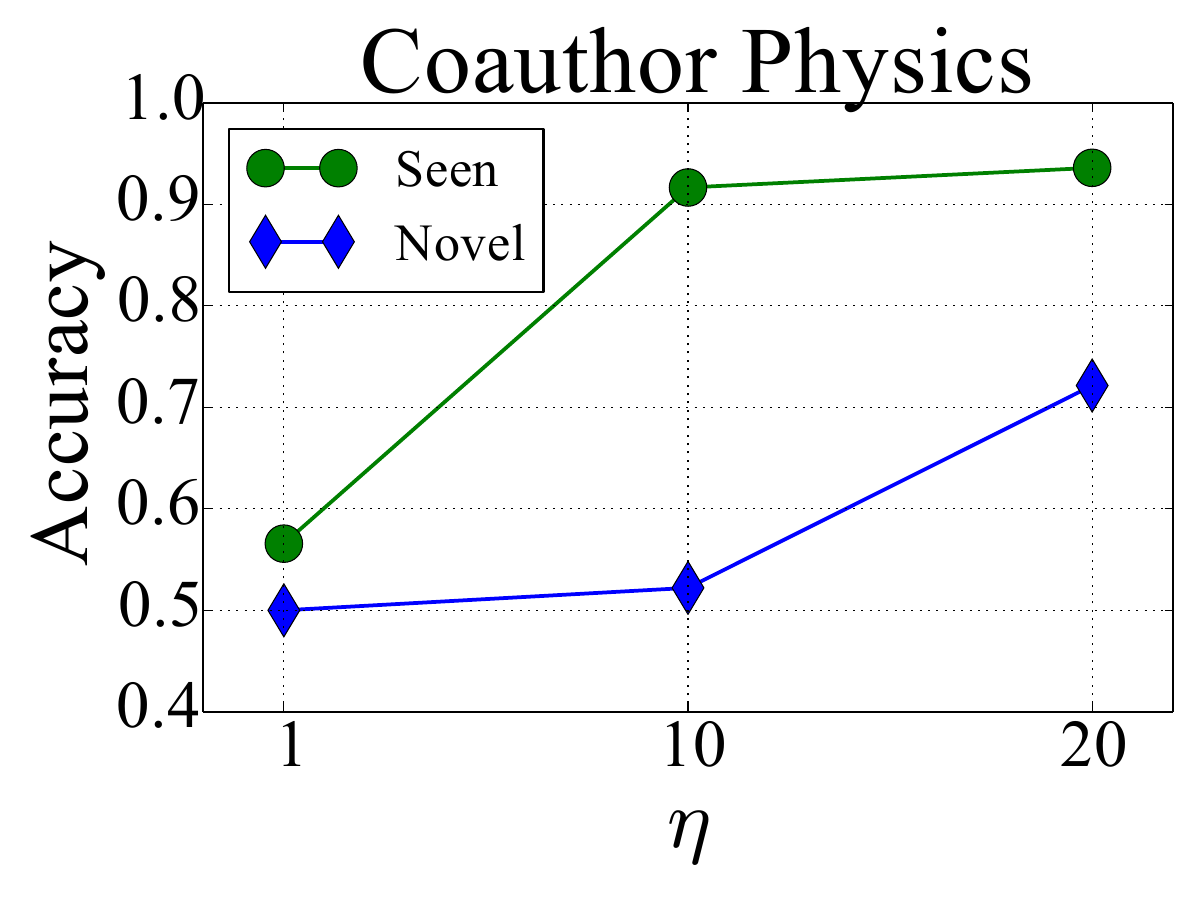}
    \includegraphics[width=0.205\textwidth]
    {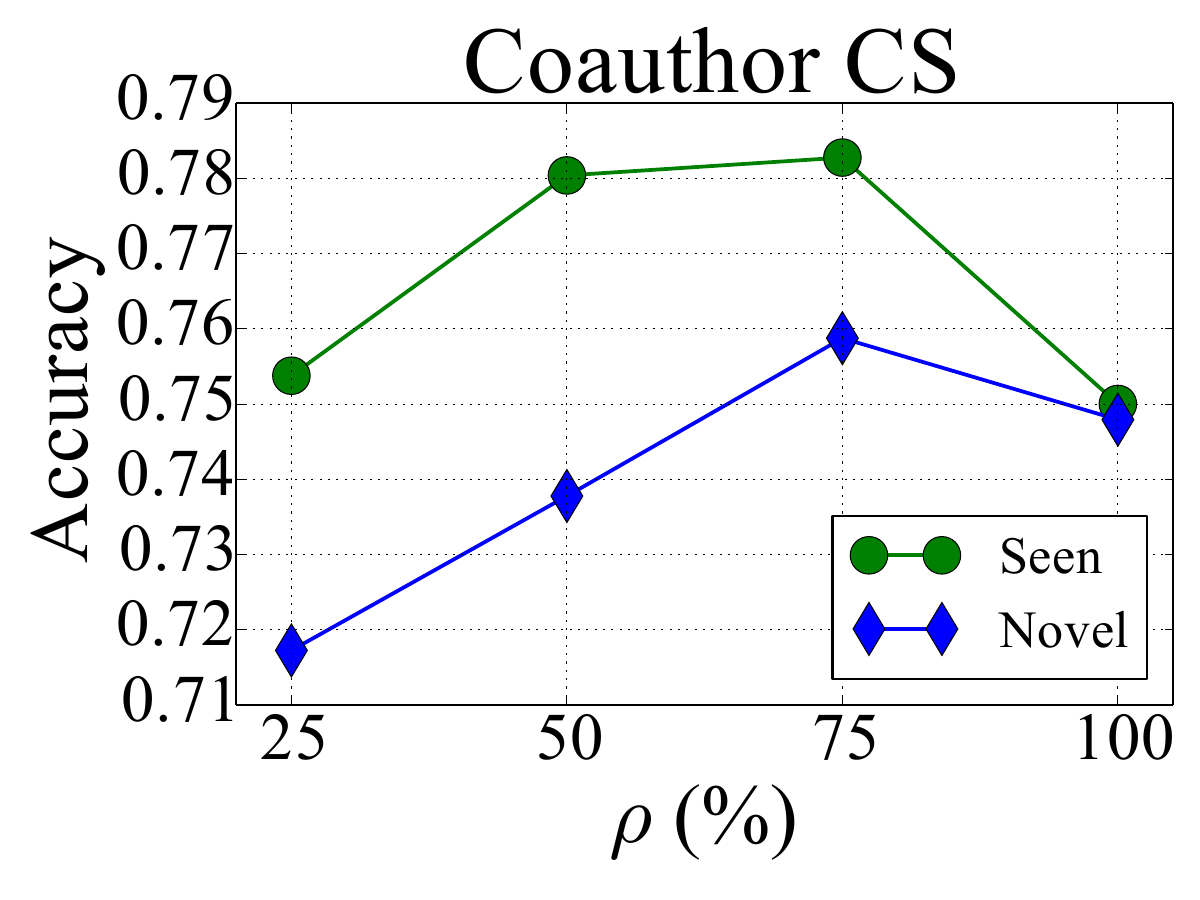}
    \hspace{0.05in}
    \includegraphics[width=0.205\textwidth]{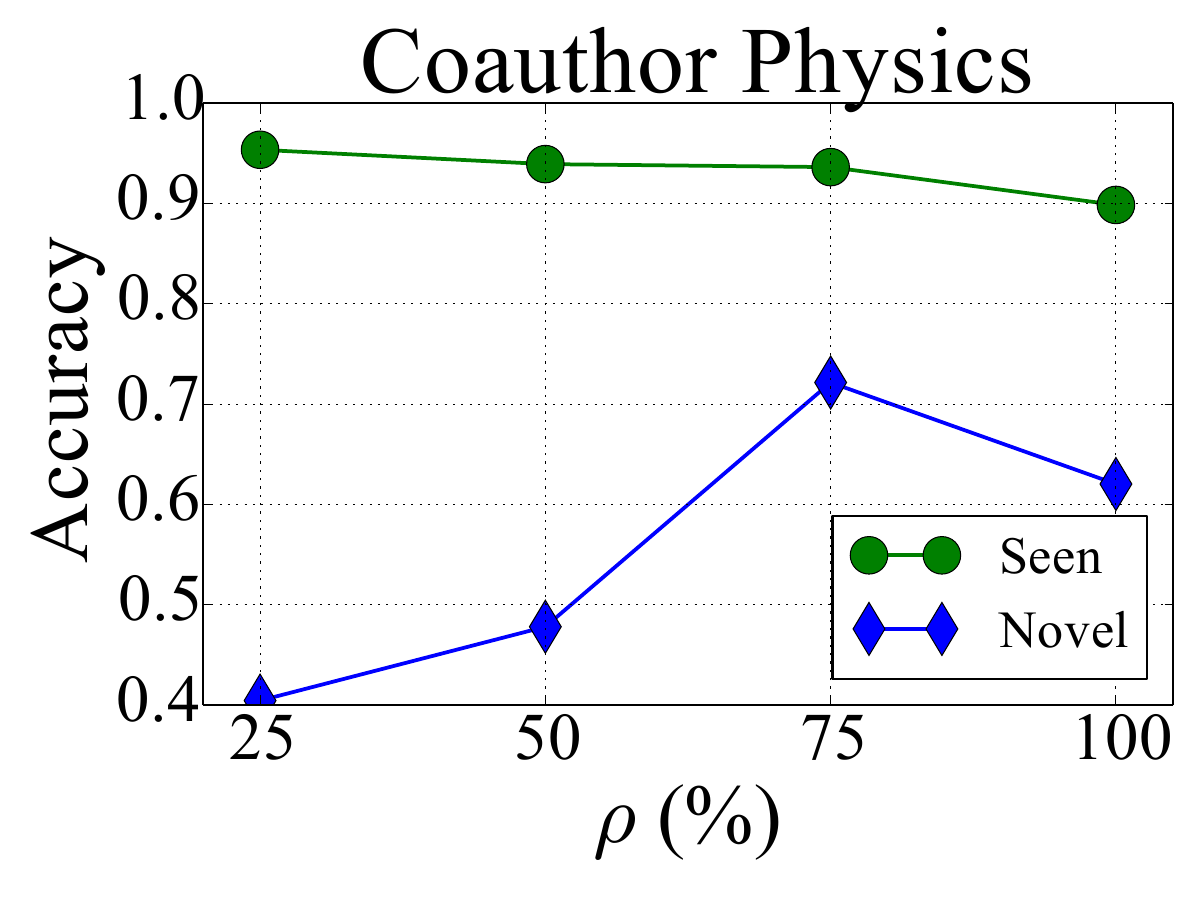}
    \caption{Effects of hyper-parameters $\eta$ and $\rho$.}
    \label{fig:BPCL_sensitivity}    
\end{figure}

\begin{table*}[ht]
  \small
  \centering
  \renewcommand\arraystretch{0.8}
  \newcolumntype{?}{!{\vrule width 1pt}}
  \caption{Overall evaluation by test accuracy without knowing the real number of novel classes (\%).}
    \label{tab:unknown_novel_class_number}
  \setlength{\tabcolsep}{1.2mm}{
    \begin{tabular}{p{2.85cm}?ccc?ccc?ccc?ccc?ccc}
    \toprule
     &\multicolumn{3}{c?}{\textbf{Citeseer}} & \multicolumn{3}{c?}{\makecell[c]{\textbf{Amazon}\\\textbf{Photos}}} & \multicolumn{3}{c?}{\makecell[c]{\textbf{Amazon}\\\textbf{Computers}}}&\multicolumn{3}{c?}{\makecell[c]{\textbf{Coauthor}\\\textbf{CS}}}
     &\multicolumn{3}{c}{\makecell[c]{\textbf{Coauthor}\\\textbf{Physics}}}\\   
    \cmidrule{2-16}
    & All & Seen & Novel& All & Seen & Novel& All & Seen & Novel& All & Seen & Novel& All & Seen & Novel\\
    \midrule
    {ORCA-ZM} & 52.2 & 70.1 & 35.1
              & 69.3 & 84.4 & 52.6
              & 66.0 & 74.3 & 57.6
              & 74.8 & 74.8 & 72.9
              & 69.7 & 63.6 & 67.5\\
    {ORCA}  & 52.8 & 65.6 & 40.2
            & 71.8 & 82.2 & 59.0 
            & 64.4 & 75.1 & 52.1
            & 72.9 & 75.6 & 70.3
            & 70.9 & 70.4 & 67.1\\
    {OpenCon} & 53.4 & 68.8 & 39.3
              & \textbf{80.9} & 92.2 & 70.3
              & 64.5 & 80.4 & 51.9
              & 76.4 & 88.8 & 66.9 
              & 58.3 & 94.9 & 44.0\\
    \midrule
    \textbf{\smodel}& \textbf{67.6} & 73.8 & 60.4 
                    & 74.7 & 77.8 & 67.4
                    & \textbf{67.0} & 72.9 & 58.2
                    & \textbf{80.2} & 78.9 & 80.0
                    & \textbf{74.4} & 72.1 & 73.9\\
    \bottomrule
   
    \end{tabular}
    }%
\end{table*}
\begin{table}
  \centering
  \small
  \renewcommand\arraystretch{0.8}
    \newcolumntype{?}{!{\vrule width 1pt}}
  \caption{Evaluation of hyper-parameter search metrics on Amazon Photos (\%). We report the absolute accuracy gap between seen and novel classes in the last column.}
    \label{tab:measure_selection}
  \setlength{\tabcolsep}{1.2mm}{
    \begin{tabular}{l?c?cccc}
    \toprule
     \multirow{2}*{\textbf{Method}} & \multirow{2}*{\textbf{Metric}} &\multicolumn{4}{c}{\textbf{Test Accuracy}}\\   
   & & All & Seen & Novel & Gap \\
   \midrule
    \multirow{3}*{ORCA-ZM} & SC & 54.4 & 67.3 & 39.0 & 28.3 \\
                           & ACC & 71.4 & 86.5 & 54.9 & 31.6 \\
                           & SC\&ACC & \textbf{74.6} & \textbf{89.9} & \textbf{58.2} & 31.7\\
    \midrule
    \multirow{3}*{ORCA} & SC & 41.4 & 44.7 & 33.9 & 10.8 \\
                        & ACC & 73.3 & 85.8 & 60.3 & 25.5 \\
                        & SC\&ACC & \textbf{76.2} & \textbf{87.1} & \textbf{64.9} & 22.2 \\
    \midrule
    \multirow{3}*{SimGCD} & SC & 79.6 & 87.7 & 71.9 & 15.8\\
                           & ACC & 79.5 & 92.1 & 66.1 & 26.0\\
                           & SC\&ACC & \textbf{80.5} & \underline{90.0} & \underline{70.8} & 19.2\\
    \midrule
    \multirow{3}*{OpenLDN} & SC & 48.6 & 48.9 & 46.0 & 2.9 \\
                           & ACC & 71.6 & 88.4 & 52.3 & 36.1 \\
                           & SC\&ACC & \textbf{80.9} & \textbf{90.6} & \textbf{71.9} & 18.7 \\
    \midrule
    \multirow{3}*{OpenCon} & SC & 83.6 & 90.8 & 76.0 & 14.8 \\
                           & ACC & 82.0 & 92.3 & 72.0 & 20.3 \\
                           & SC\&ACC & \underline{82.6} & \underline{92.1} & \underline{72.8} & 19.3 \\
    \midrule
    \multirow{3}*{OpenCon$^{\ddag}$} & SC & 80.4 & 85.7 & 74.9 & 10.8 \\
                           & ACC & 81.2 & 91.5 & 71.8 & 19.7 \\
                           & SC\&ACC & \textbf{82.9} & \underline{87.9} & \textbf{78.1} & 9.8 \\
    \midrule
    \multirow{3}*{InfoNCE} & SC & 77.0 & 77.1 & 77.5 & 0.4 \\
                           & ACC & 75.4 & 78.5 & 73.4 & 5.1 \\
                           & SC\&ACC & \underline{76.3} & \textbf{78.5} & \underline{75.1} & 3.4 \\
    \midrule
    \multirow{3}*{InfoNCE+SupCon} & SC & 77.2 & 77.5 & 77.3 & 0.2 \\
                                  & ACC & 75.5 & 79.7 & 72.4 & 7.3 \\
                                  & SC\&ACC & \underline{75.6} & \textbf{80.3} & 72.0 & 8.3 \\
    \midrule
    \multirow{3}*{InfoNCE+SupCon+CE} & SC & 77.6 & 78.5 & 77.2 & 1.3 \\
                                     & ACC & 75.5 & 79.7 & 71.8 & 7.9 \\
                                     & SC\&ACC & \underline{76.4} & \textbf{80.5} & \underline{72.9} & 7.6\\
    \midrule
    \multirow{3}*{\textbf{\smodel}} & SC & 83.3 & 89.3 & 77.1 & 12.2 \\
                                                           
                                     & ACC & 82.1 & 90.6 & 73.4 & 17.2 \\
                                     
                                     & SC\&ACC & \textbf{83.6} & \underline{89.9} & \textbf{77.3} & 12.6\\
    \bottomrule
   
    \end{tabular}
    }
\end{table}

\subsection{Effects of Hyper-Parameters}
Taking Coauthor CS/Physics as the examples,
Figure~\ref{fig:BPCL_sensitivity} shows how the scaling factor and pseudo label selection rate impact the test accuracy of \model.
(1) Scaling factor $\eta$. On Coauthor CS, the accuracy on seen classes increases as the $\eta$ increases. However, a large $\eta$ tends to suppress the accuracy on novel classes due to the over-fitting of seen classes. Notably, for Coauthor Physics, a larger $\eta$ leads to a significant gain in accuracy on seen classes (over 30\%), which suggests that the label information might be seriously underutilized if using a small $\eta$. Therefore, for the Coauthor Physics, using a larger $\eta$ improves performance on both seen and novel classes.
(2) Pseudo-label selection rate $\rho$. On Coauthor CS, properly increasing the number of pseudo labels can benefit model performance on both seen and novel classes. However, too many pseudo labels can induce more noise to suppress the model performance. Particularly on Coauthor Physics, supplementing pseudo labels inhibits the model performance on seen classes. This is because the seen classes have been well learned with a small number of pseudo labels (achieving over 90\% accuracy on seen classes with $\rho=25$). Hence increasing $\rho$ can amplify the negative impact of noisy pseudo labels of seen classes.

\subsection{Investigation of Number of Novel Classes}\label{sec:unknown_num_class}
In this section, we discuss how to deal with the situation, in which the number of novel classes is unknown.
If we have sufficient validation data, we can follow existing efforts~\cite{GCD} to estimate the number of novel classes according to clustering accuracy on the validation set before model training. 
However, each seen class could have limited label information in practice. Therefore, we treat the number of novel classes as a hyper-parameter.
In this paper, before model training, we run K-Means over node representations learned by InfoNCE, and roughly estimate the number of novel classes according to the silhouette coefficient metric. For each of the used datasest, we find that the roughly estimated number of novel classes is less than 10.  
Hence we set the number of novel classes to range from 1 to 10, and determine the value of the hyper-parameter according to the SC\&ACC metric.
In Table~\ref{tab:unknown_novel_class_number}, we compare \smodel with the most competitive baselines, including ORCA, ORCA-ZM, and OpenCon. The results show that \smodel still performs the best on most of the datasets.
This is an exploration of the unknown number of novel classes. We will further investigate how to determine the number in our future work.

\subsection{Evaluation of Hyper-Parameter Search Metrics }\label{sec:Section_eval_AC&ACC} 
Taking Amazon Photos as the example, we compare different metrics for hyper-parameter search in Table~\ref{tab:measure_selection}, and make the following observations. 
(1) Searching based on ACC (validation accuracy) tends to result in a larger accuracy gap between seen and novel classes, indicating that the resultant models are biased towards seen classes.
(2) The effectiveness of SC (silhouette coefficient) or ACC varies across different methods. ORCA-ZM, ORCA, and OpenLDN prefer ACC, while InfoNCE, InfoNCE+SupCon, and InfoNCE+SupCon+CE favor SC. In contrast, SC\&ACC has a more stable performance across various methods and datasets, better contributing to the evaluations of the open-world SSL models.
Similar observations can be found on other datasets.

\section{Proof}\label{sec:proof}
\subsection{Proof Skeleton}
In this section, we target to prove Theorem~\ref{theorem}.
We use bold lowercase letters to represent vectors, regular lowercase letters to represent scalars, and regular uppercase letters to denote matrices.
To simplify the problem, we first show that we can transform the K-Means clustering over $d$-dimension data (generated from a uniform mixture of spherical Gaussian distributions $\mathcal{N}\left(\boldsymbol{\mu_1},\Sigma_1\right)$ and $\mathcal{N}\left(\boldsymbol{\mu_2},\Sigma_2\right)$) to the K-Means clustering over 1-dimension data (generated from the uniform mixture of $\mathcal{N}\left(\mu_1,\sigma_1\right)$ and $\mathcal{N}\left(\mu_2,\sigma_2\right)$) without loss of generality.
Then we bridge the accuracy $\{ACC_1,ACC_2\}$ and the distribution parameters $\{\mu_1,\mu_2,\sigma_1,\sigma_2\}$ to prove Theorem~\ref{theorem}.

To do this, we introduce a partition threshold $s$ and 
a function $h\left(s,\mu_1,\mu_2,\sigma_1,\sigma_2\right) = 2s-\theta_1-\theta_2$. Here $\theta_1$ and $\theta_2$ are cluster centers found by K-Means based on the partition threshold $s$. Therefore, $\theta_1$ and $\theta_2$ can be represented by $\mu_1,\mu_2,\sigma_1, \sigma_2$, and $s$. 
If $(\theta_1+\theta_2)/2$ exactly equals $s$, the cluster centers will not change in the subsequent iterations of K-Means.
In other words, the optimal partition threshold $s^{*}$ is the solution of $h\left(s,\mu_1,\mu_2,\sigma_1,\sigma_2\right)=0$.
To prove the first point of Theorem~\ref{theorem}, we prove that $s^{*}$ and $\sigma_1$ have a negative correlation,
and $ACC_2$ is negatively correlated with $s^{*}$. Hence we can derive that $ACC_2$ is positively correlated to $\sigma_1$.
To prove the second point of Theorem~\ref{theorem}, we calculate value ranges of $ACC_1$ and $ACC_2$ based on value ranges of $s^{*}$, $\alpha$, and $\gamma$.

\subsection{Proof of the First Point of Theorem 1}
Given any $d$-dimension vectors $\boldsymbol{\mu_1}$ and $\boldsymbol{\mu_2}$, there exists a rotation matrix $R$ and a bias $\boldsymbol{b}$ that can transform $\boldsymbol{\mu_1}$ and $\boldsymbol{\mu_2}$ to the first dimension, i.e., 
\begin{equation}     \footnotesize
R\boldsymbol{\mu_1} + \boldsymbol{b} = \left(\mu_1,0,...,0\right),
R\boldsymbol{\mu_2} + \boldsymbol{b} = \left(\mu_2,0,...,0\right)
\end{equation}
Since K-Means clustering is based on the Euclidean distance metric, the clustering result will keep the same if all samples are rotated and moved with $R$ and $\boldsymbol{b}$.
Hence we can assume that the distribution $P_{XY}$ is symmetry about the first dimension, i.e., $\boldsymbol{\mu_1}=\left(\mu_1,0,...,0\right),\boldsymbol{\mu_2}=\left(\mu_2,0,...,0\right)$.
Here we assume that $\mu_1 < \mu_2$ and $\mu_1 = 0$.
For spherical Gaussian distributions, we can denote $\Sigma_1$ and $\Sigma_2$ by $\Sigma_1 = \textbf{I} \sigma_1^2, \Sigma_2 = \textbf{I} \sigma_2^2$. 
Without loss of generality, we assume that $\sigma_1 < \sigma_2$.

A typical method of initializing cluster centers is to average the randomly selected samples.
Since data samples are from $P_{XY}$, and $P_{XY}$ is symmetry about the first dimension, the expectations of initial cluster centers are on the first dimension, and the expectations of the cluster centers predicted by each iteration of K-Means are also on the first dimension. Therefore, the expectations of final optimal cluster centers also locate on the first dimension, represented by $\boldsymbol{\theta_1}^{*} = \left(\theta_1^*,0,...,0\right),\boldsymbol{\theta_2}^{*} = \left(\theta_2^*,0,...,0\right)$. Up to now, we have transformed the K-Means on a $d$-dimension distribution to the K-Means on a 1-dimension distribution without loss of generality.

Suppose we have a $d$-dimensional variable $\left(x_1,x_2,...,x_d\right)$. According to the property of marginal distribution in multivariate spherical Gaussian distribution, the probability density function of 
$x_1$ is $\mathbbm{P}\left(x_1 = t\right) = \frac{1}{2}\mathbbm{P}\left[\mathcal{N}\left(\mu_1,\sigma_1\right)=t\right] + \frac{1}{2}\mathbbm{P}\left[\mathcal{N}\left({\mu_2},\sigma_2\right)=t\right]$. Given the partition threshold $s$ found by K-Means, we can calculate the expectation of current cluster center $\boldsymbol{\theta_1}=\left(\theta_1,0,...,0\right)$ by
\begin{equation}     
\footnotesize
\label{expectation of current cluster centers}
\begin{aligned}
  \theta_1 &= \mathbbm{E}\left[x_1 | x_1 < s\right] = \frac{\mathbbm{E}\left[x_1 \mathbbm{1}\left(x_1 < s\right)\right]}{\mathbbm{P}\left(x_1 < s\right)}  \\
  &=\frac{\frac{1}{2}\mathbbm{E}_{x_1\sim \mathcal{N}\left(\mu_1,\sigma_1\right)}\left[x_1 | x_1 < s\right]+\frac{1}{2}\mathbbm{E}_{x_1\sim \mathcal{N}\left(\mu_2,\sigma_2\right)}\left[x_1 | x_1 < s\right]}{\frac{1}{2}\mathbbm{P}_{x_1\sim \mathcal{N}\left(\mu_1,\sigma_1\right)}\left(x_1 < s\right)+\frac{1}{2}\mathbbm{P}_{x_1\sim \mathcal{N}\left(\mu_2,\sigma_2\right)}\left(x_1 < s\right)}
\end{aligned}
\end{equation}

Let $\Phi\left(x\right)$ and $\varphi\left(x\right)$ be the cumulative distribution function (cdf) and probability density function (pdf) of the standard normal distribution, respectively, we have
\begin{equation}     \footnotesize
    \mathbbm{P}_{x_1\sim \mathcal{N}\left(\mu_1,\sigma_1\right)}\left(x_1 < s\right) = \Phi\left(\frac{s-\mu_1}{\sigma_1}\right)
\end{equation}
\begin{equation}     \footnotesize
    \mathbbm{P}_{x_1\sim \mathcal{N}\left(\mu_2,\sigma_2\right)}\left(x_1 < s\right) = \Phi\left(\frac{s-\mu_2}{\sigma_2}\right)
\end{equation}

According to Lemma~\ref{lemma1} in Section~\ref{sec:lemma_1}, we have
\begin{equation}     \footnotesize
    \mathbbm{E}_{x_1\sim \mathcal{N}\left(\mu_1,\sigma_1\right)}\left[x_1 | x_1 < s\right] = \mu_1 \Phi\left(\frac{s-\mu_1}{\sigma_1}\right) - \sigma_1 \varphi\left(\frac{s-\mu_1}{\sigma_1}\right)
\end{equation}
\begin{equation}     \footnotesize
    \mathbbm{E}_{x_1\sim \mathcal{N}\left(\mu_2,\sigma_2\right)}\left[x_1 | x_1 < s\right] = 
\mu_2 \Phi\left(\frac{s-\mu_2}{\sigma_2}\right) - \sigma_2 \varphi\left(\frac{s-\mu_2}{\sigma_2}\right)
\end{equation}

Substituting the above equations into Eq.~\ref{expectation of current cluster centers}, we can derive
\begin{equation}     \footnotesize
    \theta_1 = \frac{\mu_1 \Phi\left(\frac{s-\mu_1}{\sigma_1}\right) - \sigma_1 \varphi\left(\frac{s-\mu_1}{\sigma_1}\right) + \mu_2 \Phi\left(\frac{s-\mu_2}{\sigma_2}\right) - \sigma_2 \varphi\left(\frac{s-\mu_2}{\sigma_2}\right)}{\Phi\left(\frac{s-\mu_1}{\sigma_1}\right) + \Phi\left(\frac{s-\mu_2}{\sigma_2}\right)}
    \label{eq:theta_1}
\end{equation}

Similarly, we can derive
\begin{equation}     
    \footnotesize
    \theta_2 = \frac{\mu_1 - \mu_1 \Phi\left(\frac{s-\mu_1}{\sigma_1}\right) + \sigma_1 \varphi\left(\frac{s-\mu_1}{\sigma_1}\right) + \mu_2 - \mu_2 \Phi\left(\frac{s-\mu_2}{\sigma_2}\right) + \sigma_2 \varphi\left(\frac{s-\mu_2}{\sigma_2}\right)}{1 - \Phi\left(\frac{s-\mu_1}{\sigma_1}\right) + 1 - \Phi\left(\frac{s-\mu_2}{\sigma_2}\right)}
    \label{eq:theta_2}
\end{equation}

According to Eq.~\ref{eq:theta_1} and Eq.~\ref{eq:theta_2}, as $\theta_1$ and $\theta_2$ can be represented by $\mu_1,\mu_2,\sigma_1, \sigma_2$, and $s$, we define a function $h\left(s,\sigma_1,\sigma_2,\mu_1,\mu_2\right) = 2s-\theta_1-\theta_2$. Notably, $\theta_1$ and $\theta_2$ are determined based on the current $s$.
If $s$ exactly equals $(\theta_1+\theta_2)/2$, the cluster centers will not change in the subsequent iterations of K-Means.
In other words, the optimal partition threshold $s^{*}$ found by K-Means is the solution of $h\left(s,\sigma_1,\sigma_2,\mu_1,\mu_2\right) = 0$.

Now we target to discuss the monotonicity of functions $h\left(s\right)$ and $h\left(\sigma_1\right)$, aiming to explore the correlation between $\sigma_1$ and $ACC_2$.
Before calculating the derivatives, we restrict $s$ into a smaller interval to simplify our proof. As proved in previous studies, the optimal cluster centers found by K-Means can be very close to the ground-truth cluster centers~\cite{kalai2010efficiently}, that is
\begin{equation}     
    \footnotesize
    \label{error range of optimal centers}
    | \theta^{*}_1 - \mu_1 | < \epsilon, | \theta^{*}_2 - \mu_2 | < \epsilon
\end{equation}

\noindent where $\epsilon$ is a rather small value (compared to $\mu_2-\mu_1,\sigma_1, \sigma_2$). 
Thus, we have
\begin{equation} 
    \footnotesize
    \label{error range of optimal threashold}
    |s^{*} - \frac{\mu_1 +\mu_2}{2}| = |\frac{\theta_1^{*} +\theta_2^{*}}{2} - \frac{\mu_1 +\mu_2}{2}| \leq \frac{|\theta_1^{*} - \mu_1|}{2} + \frac{|\theta_2^{*} - \mu_2|}{2} < \epsilon
\end{equation}

That is to say, $s^{*} \in \left[\frac{\mu_1 +\mu_2}{2} - \epsilon,\frac{\mu_1 +\mu_2}{2} + \epsilon \right]$.
We will prove that $h\left(s\right)$ is a strictly monotone increasing function within the interval $\left[\frac{\mu_1 +\mu_2}{2} - \epsilon,\frac{\mu_1 +\mu_2}{2} + \epsilon \right]$.

For the second part of $h\left(s\right)$, we calculate the derivative
\begin{equation} 
\footnotesize
\label{eq_c1-d}
\begin{aligned}
\frac{\partial \theta_1}{\partial s}
&=\frac{\frac{\mu_2-\mu_1}{\sigma_2}\varphi\left(\frac{s-\mu_2}{\sigma_2}\right)\Phi\left(\frac{s-\mu_1}{\sigma_1}\right) - \frac{\mu_2-\mu_1}{\sigma_1}\varphi\left(\frac{s-\mu_1}{\sigma_1}\right)\Phi\left(\frac{s-\mu_2}{\sigma_2}\right)}{\left(\Phi\left(\frac{s-\mu_1}{\sigma_1}\right) + \Phi\left(\frac{s-\mu_2}{\sigma_2}\right)\right)^2}\\
&\quad + \frac{\frac{s-\mu_1}{\sigma_1}\varphi\left(\frac{s-\mu_1}{\sigma_1}\right)+\frac{s-\mu_2}{\sigma_2}\varphi\left(\frac{s-\mu_2}{\sigma_2}\right)}{\Phi\left(\frac{s-\mu_1}{\sigma_1}\right) + \Phi\left(\frac{s-\mu_2}{\sigma_2}\right)}\\
&\quad + \frac{\left(\sigma_1 \varphi\left(\frac{s-\mu_1}{\sigma_1}\right)+\sigma_2 \varphi\left(\frac{s-\mu_2}{\sigma_2}\right)\right)\left(\frac{\varphi\left(\frac{s-\mu_1}{\sigma_1}\right)}{\sigma_1}+\frac{\varphi\left(\frac{s-\mu_2}{\sigma_2}\right)}{\sigma_2}\right)}{\left(\Phi\left(\frac{s-\mu_1}{\sigma_1}\right) + \Phi\left(\frac{s-\mu_2}{\sigma_2}\right)\right)^2}
\end{aligned}
\end{equation}

With the restriction of $\alpha$-separated (Cf. Definition \ref{def:separate}), we have $||\boldsymbol{\mu_1}-\boldsymbol{\mu_2} ||_2 = \alpha\left(\sigma_1 + \sigma_2\right)$. Since we have assumed that $\mu_2 > \mu_1$, we can derive
\begin{equation} 
    \footnotesize
    \mu_2 - \mu_1 = \alpha \left(\sigma_1+\sigma_2\right) = \alpha \left(1+\gamma\right) \sigma_1 = \alpha \left(1+1/\gamma\right) \sigma_2
\end{equation}

As $1.5 <\alpha < 3, 1<\gamma <2 $, we have
\begin{equation}     \footnotesize
3 \times \left(1+1 \right) > \frac{\mu_2 - \mu_1}{\sigma_2} = \alpha \left(1+1/ \gamma\right) > 1.5 \times \left(1+0.5\right) = 2.25
\end{equation}
\begin{equation}
 \footnotesize
    3 \times \left(1+2 \right) > \frac{\mu_2 - \mu_1}{\sigma_1} = \alpha \left(1+ \gamma\right) > 1.5 \times \left( 1+ 2 \right) = 3  
\end{equation}

Then, we can obtain the inequalities below,
\begin{equation}
    \footnotesize
    \begin{aligned}
    \varphi\left(\frac{s-\mu_2}{\sigma_2}\right) &< \varphi\left(\frac{\left(\frac{\mu_1 +\mu_2}{2} + \epsilon\right)-\mu_2}{ \sigma_2}\right) \\
    &= \varphi\left(\frac{\mu_2-\mu_1}{2 \sigma_2}-\frac{\epsilon}{\sigma_2} \right) 
 < \varphi\left(1.1\right) 
    \end{aligned}
\end{equation}
\begin{equation}     
    \footnotesize
    \begin{aligned}
    \varphi\left(\frac{s-\mu_1}{\sigma_1}\right) &< \varphi\left(\frac{\left(\frac{\mu_1 +\mu_2}{2} - \epsilon\right)-\mu_1}{\sigma_1}\right) \\
    &= \varphi\left(\frac{\mu_2-\mu_1}{2 \sigma_1}-\frac{\epsilon}{\sigma_1} \right) 
 < \varphi\left(1.45\right) 
    \end{aligned}
\end{equation}
\begin{equation}     \footnotesize
    \Phi\left(1.5\right)  < \Phi\left(\frac{s-\mu_1}{\sigma_1}\right) < \Phi\left(3\right),  \Phi\left(\frac{s-\mu_2}{\sigma_2}\right) < 1 - \Phi\left(1.1\right)  
\end{equation}

According to the pdf values and cdf values of the standard normal distribution, for the first part of Eq.~\ref{eq_c1-d}, we have
\begin{equation}     \footnotesize
    \begin{aligned}
    &\frac{\frac{\mu_2-\mu_1}{\sigma_2}\varphi\left(\frac{s-\mu_2}{\sigma_2}\right)\Phi\left(\frac{s-\mu_1}{\sigma_1}\right) - \frac{\mu_2-\mu_1}{\sigma_1}\varphi\left(\frac{s-\mu_1}{\sigma_1}\right)\Phi\left(\frac{s-\mu_2}{\sigma_2}\right)}{\left(\Phi\left(\frac{s-\mu_1}{\sigma_1}\right) + \Phi\left(\frac{s-\mu_2}{\sigma_2}\right)\right)^2}<  0.8
    \end{aligned}
\end{equation}

For the second part of Eq.~\ref{eq_c1-d}, we have\footnote{As the left side of Eq.~\ref{eq:numerical_scaling} is not a monotonic function, we use numerical simulation to determine the threshold.}
\begin{equation}     \footnotesize
    \begin{aligned}
    \frac{\frac{s-\mu_1}{\sigma_1}\varphi\left(\frac{s-\mu_1}{\sigma_1}\right)+\frac{s-\mu_2}{\sigma_2}\varphi\left(\frac{s-\mu_2}{\sigma_2}\right)}{\Phi\left(\frac{s-\mu_1}{\sigma_1}\right) + \Phi\left(\frac{s-\mu_2}{\sigma_2}\right)} < 0.1
    \end{aligned}
    \label{eq:numerical_scaling}
\end{equation}

For the third part of Eq.~\ref{eq_c1-d}, we have
\begin{equation}     \footnotesize
    \begin{aligned}
    &\frac{\left(\sigma_1 \varphi\left(\frac{s-\mu_1}{\sigma_1}\right)+\sigma_2 \varphi\left(\frac{s-\mu_2}{\sigma_2}\right)\right)\left(\frac{\varphi\left(\frac{s-\mu_1}{\sigma_1}\right)}{\sigma_1}+\frac{\varphi\left(\frac{s-\mu_2}{\sigma_2}\right)}{\sigma_2}\right)}{\left(\Phi\left(\frac{s-\mu_1}{\sigma_1}\right) + \Phi\left(\frac{s-\mu_2}{\sigma_2}\right)\right)^2}
    < 0.1
    \end{aligned}
\end{equation}

Summing up the above inequalities, we derive $\frac{\partial \theta_1}{\partial s} < 1$, where $\mu_1 < \mu_2$, and $\sigma_1 < \sigma_2$.
As the relative size relationship between $\sigma_1$ and $\sigma_2$ does not affect the above proof, and $-\mu_1 > -\mu_2$, we have $\frac{\partial -\theta_2}{\partial -s} < 1$, and thus we can derive $\frac{\partial \theta_2}{\partial s} < 1$.
Therefore, we have
\begin{equation}     \footnotesize
    \frac{\partial h\left(s\right)}{\partial s} = 2 - \frac{\partial \theta_1}{\partial s} - \frac{\partial \theta_2}{\partial s} > 2 -1 -1 = 0
\end{equation}

Up to now, we have proved that $h\left(s\right)$ is strictly monotone increasing within $\left[\frac{\mu_1 +\mu_2}{2} - \epsilon,\frac{\mu_1 +\mu_2}{2} + \epsilon\right]$.

Next, we discuss the monotonicity of $h\left(\sigma_1\right)$ by calculating the derivative below,
\begin{equation}     \footnotesize
    \frac{\partial h\left(\sigma_1\right)}{\partial \sigma_1} = -\frac{\partial \theta_1}{\partial \sigma_1} -\frac{\partial \theta_2}{\partial \sigma_1}
\end{equation}

We have assumed that $\mu_1 = 0$, so that we can derive
\begin{equation}     
    \footnotesize
   \label{sigma_d} 
   \begin{aligned}
-\frac{\partial \theta_1}{\partial \sigma_1} &= 
\frac{\left(1 +  \frac{\left(s-\mu_1\right)^2}{\sigma_1^2}\right)\varphi\left(\frac{s-\mu_1}{\sigma_1}\right)\left(\Phi\left(\frac{s-\mu_1}{\sigma_1}\right)+\Phi\left(\frac{s-\mu_2}{\sigma_2}\right)\right) }{\left(\Phi\left(\frac{s-\mu_1}{\sigma_1}\right)+\Phi\left(\frac{s-\mu_2}{\sigma_2}\right)\right)^2}\\
&+\frac{\left[\sigma_1 \varphi\left(\frac{s-\mu_1}{\sigma_1}\right) - \mu_2 \Phi\left(\frac{s-\mu_2}{\sigma_2}\right) + \sigma_2 \varphi\left(\frac{s-\mu_2}{\sigma_2}\right)\right]\frac{s-\mu_1}{\sigma_1^2}\varphi\left(\frac{s-\mu_1}{\sigma_1}\right)}{\left(\Phi\left(\frac{s-\mu_1}{\sigma_1}\right)+\Phi\left(\frac{s-\mu_2}{\sigma_2}\right)\right)^2} 
\end{aligned}
\end{equation}

The denominator of Eq.~\ref{sigma_d} is positive. For the only negative term in the numerator of Eq.~\ref{sigma_d}, we can obtain that
\begin{equation}     \footnotesize
    \begin{aligned}
    &\mu_2 \Phi\left(\frac{s-\mu_2}{\sigma_2}\right)\frac{s-\mu_1}{\sigma_1^2}\varphi\left(\frac{s-\mu_1}{\sigma_1}\right)\\
    &=\left(\mu_2-\mu_1\right) \Phi\left(\frac{s-\mu_2}{\sigma_2}\right)\frac{s-\mu_1}{\sigma_1^2}\varphi\left(\frac{s-\mu_1}{\sigma_1}\right)\\
    &<\frac{1}{2}\left(\mu_2-\mu_1\right)\frac{s-\mu_1}{\sigma_1^2}\varphi\left(\frac{s-\mu_1}{\sigma_1}\right)\left(\Phi\left(\frac{s-\mu_1}{\sigma_1}\right)+\Phi\left(\frac{s-\mu_2}{\sigma_2}\right)\right)\\
     &< \left(1 +  \frac{\left(s-\mu_1\right)^2}{\sigma_1^2}\right)\varphi\left(\frac{s-\mu_1}{\sigma_1}\right)\left(\Phi\left(\frac{s-\mu_1}{\sigma_1}\right)+\Phi\left(\frac{s-\mu_2}{\sigma_2}\right)\right)
    \end{aligned}
\end{equation}
That is to say, the only negative term in the numerator is smaller than one of the positive term in the numerator. Therefore, we have $-\frac{\partial \theta_1}{\partial \sigma_1} > 0$. 
$\theta_1$ and $\theta_2$ are determined by the K-Means algorithm.
Notably, we have transformed to conduct the K-Means on a 1-dimension
distribution, and assume that $\mu_1 < \mu_2$. 
As the partition threshold $s$ increases, $\theta_2$ will increase, and $\theta_1$ will also increase because more larger instances are assigned to the first cluster. 
As $s$ decreases, $\theta_1$ will decrease, and $\theta_2$ will also decrease because more smaller instances are assigned to the second cluster. In other words, $\theta_1$ and $\theta_2$ are positively correlated. 
Hence we derive that $-\frac{\partial \theta_2}{\partial \sigma_1} > 0$, and thus we have $\frac{\partial h\left(\sigma_1\right)}{\partial \sigma_1} > 0$, i.e., $h\left(\sigma_1\right)$ is strictly monotone increasing.

We have proved that both $h\left(s\right)$ and $h\left(\sigma_1\right)$ are strictly monotone increasing functions.
Now we aim to discuss the correlation between $\sigma_1$ and $ACC_2$ based on the above proved conclusions.
Firstly, we formulate $ACC_2$ as follows,
\begin{equation}     \footnotesize
    \label{ACC2 formulation}
    \begin{aligned}
ACC_2 =& \mathbbm{E}\left[\mathbbm{1}\left(\hat{Y}=2\right)|Y=2\right]=\mathbbm{E}\left[\mathbbm{1}\left(x_1>s\right)|Y=2\right] \\
=&\mathbbm{P}_{x_1 \sim \mathcal{N}\left(\mu_2,\sigma_2\right)}\left(x_1>s\right)=1-\Phi\left(\frac{s-\mu_2}{\sigma_2}\right)        
    \end{aligned}
\end{equation}

We prove $ACC_2$ and $\sigma_1$ a.s. are positively correlated by:
\begin{itemize}[leftmargin=1em]
    \item Given two candidate values $\sigma_1^{\prime}$ and $\sigma_1^{\prime\prime}$ for $\sigma_1$, suppose that $\sigma_1^{\prime} < \sigma_1^{\prime\prime}$, we have $h_{\sigma_1^{\prime}}\left(s\right)<h_{\sigma_1^{\prime\prime}}\left(s\right)$ because $h\left(\sigma_1\right)$ is strictly monotone increasing.
    
    \item Suppose that $s^{*\prime}$ is the solution of $h_{\sigma_1^{\prime}}\left(s\right)=0$, and $s^{*\prime\prime}$ is the solution for $h_{\sigma_1^{\prime\prime}}\left(s\right)=0$, we have $h_{\sigma_1^{\prime\prime}}\left(s^{*\prime}\right)>h_{\sigma_1^{\prime}}\left(s^{*\prime}\right)=h_{\sigma_1^{\prime\prime}}\left(s^{*\prime\prime}\right)=0$. Considering that $h\left(s\right)$ is  a strictly monotone increasing function, we have $s^{*\prime} > s^{*\prime\prime}$.
    Thus, if $\sigma_1^{\prime} < \sigma_{1}^{\prime\prime}$, we have $s^{*\prime} > s^{*\prime\prime}$. In other words, $s^{*}$ and $\sigma_1$ have a negative correlation.

    \item Substituting $s^{*\prime}$ and $s^{*\prime\prime}$ into Eq.~\ref{ACC2 formulation} ($s^{*\prime} > s^{*\prime\prime}$), we have $1-\Phi\left(\frac{s^{*\prime}-\mu_2}{\sigma_2}\right) < 1-\Phi\left(\frac{s^{*\prime\prime}-\mu_2}{\sigma_2}\right)$. That is to say, $ACC_2$ and $s^{*}$ also has a negative correlation.

    \item  As we have proved (1) $s^{*}$ and $\sigma_1$ have a negative correlation, and (2) $ACC_2$ and $s^{*}$ also have a negative correlation, we finally prove that $ACC_2$ is positively correlated to $\sigma_1$.
\end{itemize}

According to the law of large numbers and Chebyshev inequality, if we perform sampling a large number of times, i.e., generating a large amount of data samples according to $P_{XY}$, the calculated results should be very close to the expectations. Therefore, 
there exits a constant $\overline{N}$, if the number of data samples $N \geq \overline{N}$, with a possibility at least 1-$\delta$, $ACC_2$ and $\sigma_1$ a.s. have a positive correlation.

\subsection{Proof of the Second Point of Theorem 1}
As mentioned before, the optimal cluster centers found by K-Means are very close to the ground-truth cluster centers~\cite{kalai2010efficiently}, so we can derive $s^{*} \in \left[\frac{\mu_1 +\mu_2}{2} - \epsilon,\frac{\mu_1 +\mu_2}{2} + \epsilon \right]$ according to Eq.~\ref{error range of optimal centers} and Eq.~\ref{error range of optimal threashold}.
Here we relax the value range to be $ s^{*}\in \left[\frac{\mu_1+\mu_2}{2} - \frac{\sigma_1}{2},\frac{\mu_1+\mu_2}{2} + \frac{\sigma_2}{2}\right]$.
Based on the relaxed range of $s^{*}$ and the restriction that $\alpha > 3$, we have
\begin{equation}     
    \footnotesize
    \begin{aligned}
    ACC_1 &= \Phi\left(\frac{s-\mu_1}{\sigma_1}\right) >  \Phi\left(\frac{\frac{\mu_1+\mu_2-\sigma_1}{2} -\mu_1}{\sigma_1}\right) = \Phi\left(\frac{\mu_2-\mu_1}{2\sigma_1} - \frac{1}{2}\right) \\
        &=\Phi\left(\frac{\alpha\left(1+ \gamma\right)-1}{2}\right) > \Phi\left(2.5\right) > 0.99
    \end{aligned}
\end{equation}
\begin{equation}     
    \footnotesize
    \begin{aligned}
    ACC_2 &= 1- \Phi\left(\frac{s-\mu_2}{\sigma_2}\right) > 1 - \Phi\left(\frac{\frac{\mu_1+\mu_2+\sigma_2}{2} -\mu_2}{\sigma_2}\right) \\
    &= 1 - \Phi\left(\frac{\mu_1-\mu_2}{2\sigma_2} + \frac{1}{2}\right) \\
        &=\Phi\left(\frac{\alpha\left(1+ 1 / \gamma\right)-1}{2}\right) > \Phi\left(1.75\right) > 0.95
    \end{aligned}
\end{equation}

Thus, we derive $|1- ACC_1|<0.05$ and $|1- ACC_2|<0.05$.

\subsection{Lemma 1 and Its Proof}\label{sec:lemma_1}
\begin{lemma}
    \label{lemma1}
    For $X\sim \mathcal{N}\left(\mu,\sigma^2\right)$, the expectation of truncated normal distribution is
    \begin{equation}     \footnotesize
        \mathbbm{E}_{x\sim \mathcal{N}\left(\mu,\sigma^2\right)}\left[x | a < x < b\right] = \mu\left[\Phi\left(\beta\right)-\Phi\left(\alpha\right)\right]-\sigma\left[\varphi\left(\beta\right)-\varphi\left(\alpha\right)\right]
    \end{equation}
where $\alpha = \frac{a-\mu}{\sigma}$ and $\beta = \frac{b-\mu}{\sigma}$.
\end{lemma}

\textbf{Proof:} Converting the expectation into integral form to have
\begin{equation}     \footnotesize
\begin{aligned}
&\mathbbm{E}_{x\sim \mathcal{N}\left(\mu,\sigma\right)}\left[x | a < x < b\right]
=\int_{a}^{b} \frac{x}{\sqrt{2\pi}\sigma} e^{-\frac{\left(x-\mu\right)^2}{2\sigma^2}}dx \\
&=\int^{\frac{b-\mu}{\sigma}}_{\frac{a-\mu}{\sigma}} \frac{\left(\sigma z + \mu\right) }{\sqrt{2\pi}}e^{-\frac{z^2}{2}}dz\\
&=\frac{\sigma}{\sqrt{2\pi}} \int^{\frac{b-\mu}{\sigma}}_{\frac{a-\mu}{\sigma}} z e^{-\frac{z^2}{2}}dz + \mu \int^{\frac{b-\mu}{\sigma}}_{\frac{a-\mu}{\sigma}} \frac{e^{-\frac{z^2}{2}}}{\sqrt{2\pi}}dz\\
&=-\sigma \frac{e^{-\frac{z^2}{2}}}{\sqrt{2\pi}}\bigg|^{\frac{b-\mu}{\sigma}}_{\frac{a-\mu}{\sigma}} + \mu \left[\Phi\left(\frac{b-\mu}{\sigma}\right)-\Phi\left(\frac{a-\mu}{\sigma}\right)\right]\\
&=\mu\left[\Phi\left(\beta\right)-\Phi\left(\alpha\right)\right]-\sigma\left[\varphi\left(\beta\right)-\varphi\left(\alpha\right)\right]\\
\end{aligned} 
\end{equation}

\section{Detailed Experimental Settings}\label{sec:appendix_hyperparam_setting}
For all models, we use the graph attention network GAT~\cite{GAT} as the feature encoder and the Adam optimizer with weight decay 1e-4 for optimization.
We set the number of GAT layers to 2, the hidden dimension to 128, the number of attention heads to 8, and the dropout rate~\cite{dropout} to 0.5.
Due to the dropout strategy, we follow SimCSE~\cite{danqi2021@SimCSE} to pass the same input to the encoder twice to obtain the positive pairs for CL.

\textbf{Settings of maximal training epochs and batch size.} We set the maximal training epochs (max\_epoch) to 100 for end-to-end models and 20 for two-stage models. Specially, we set the max\_epoch to 50 for ORCA, ORCA-ZM, and SimGCD, as we find that max\_epoch=50 would be better for the three baselines.
For ogbn-Products that has millions of nodes, a model can converge within a smaller number of training epochs because the dataset can be partitioned into more mini-batches during a training epoch. Therefore, we set max\_epoch to 10 for all methods on ogbn-Products.
For methods that adopt mini-batch training, we set the batch size to 2048 (4096 on the larger datasets ogbn-Arxiv and ogbn-Products).

\textbf{Settings of model-specific parameters.} For the baselines, their specific hyper-parameters are set to the empirical values given by the authors.
For \model, we set the scaling factor $\eta$ to 1, the temperature parameter $\tau$ to 0.7, and the pseudo-label selection rate $\rho \left(\%\right)$ to 75 by default.
Specially, we observe that properly increasing the value of $\eta$ can significantly boost the validation accuracy on Amazon Computers and Coauthor Physics (accuracy gains on the validation sets $>$ 15\%).
Hence we search $\eta\in\{10, 20\}$ for the two datasets. 
On Amazon Photos, we find that the end-to-end baselines tend to perform better than the two-stage baselines in overall accuracy. That is to say, the classification head plays an important role on Amazon Photos, so we also search $\eta\in\{10, 20\}$ for this dataset.
On Amazon Photos, Amazon Computers, and Coauthor Physics, increasing $\eta$ benefits the model performance, making the pseudo labels more reliable (especially for pseudo labels of seen classes), so we use a smaller temperature parameter $\tau=0.07$ to obtain more confident contrastive predictions.
Besides, we observe that the validation accuracy on Citeseer and ogbn-Arxiv is noticeably lower than that on other datasets, indicating a higher learning difficulty on the two datasets. Hence we lower $\rho (\%)$ to 25 on Citeseer and ogbn-Arxiv to reduce the noisy pseudo labels.

\textbf{Settings of learning rate.} We observe that the optimal learning rates of end-to-end models tend to be different on the same dataset, so we search the learning rate for each end-to-end model.
The CL-based two-stage models (i.e., InfoNCE, InfoNCE+SupCon, InfoNCE+SupCon+CE, and \model) share the similar learning scheme, and we observe that they tend to share the same optimal learning rate on the same dataset. 
Intuitively, if instances have rich initial features, it could be easier for CL to capture the semantic relationships between them, and thus a smaller learning rate is preferred for a stable convergence.
According to this, we set the default learning rate of the two-stage CL-based models to 1e-4 on Coauthor CS and Coauthor Physics, 1e-3 on Citeseer, Amazon Photos, and Amazon Computers, and 1e-2 on ogbn-Arxiv and ogbn-Products.
As a larger $\eta$ is used on Amazon Photos, Amazon Computers, and Coauthor Physics, the gradient produced by CE will be strengthened, so we increase the learning rate of \smodel by an order of magnitude on the three datasets to strengthen the gradients contributed by BPCL.
On Coauthor Physics, we set the learning rate to 1e-2 for the CL-based two-stage baselines.
We test other settings of learning rate for all the CL-based two-stage baselines and find that the above setting is the optimal choice.

\section{Conclusion}
This work is one pioneer study of open-world SSL for node classification.
Due to an absence of general pre-trained graph encoders, it is challenging to simultaneously learn effective node representations for both labeled seen classes and unlabeled novel classes, resulting in an imbalance of intra-class variances between seen and novel classes.
Based on the empirical and theoretical analysis, we propose a simple and effective method named \model, which alleviates the variance imbalance issue without relying on general pre-trained graph encoders.
Extensive experiments on the widely used graph datasets show the effectiveness of \model.
We hope this effort could be inspiring for future work on open-world graph learning.

\section*{Acknowledgment}
This work is supported by the National Key Research \& Develop Plan (2023YFF0725100) and the National Natural Science Foundation of China (62322214, U23A20299, 62076245, 62072460, 62172424, 62276270).
This work is supported by the Australian Research Council under the streams of Future Fellowship (Grant No.FT210100624) and the Discovery Project (Grant No.DP240101108). 
We also acknowledge the support from the Public Computing Cloud, Renmin University of China.
We sincerely thank the reviewers for their valuable comments.

\clearpage
\balance
\bibliographystyle{IEEEtran}
\bibliography{sample-base}

\end{document}